\documentclass{article}
\usepackage{amsmath}
\usepackage{booktabs}
\usepackage{colortbl}
\usepackage{multirow}
\usepackage{xspace}
\usepackage{tikz}
\usetikzlibrary{positioning,arrows.meta,fit,calc}
\usepackage[onecolumn]{trustai4s-template/trustai4s}
\usepackage{hyperref}                %
\usepackage{xurl}

\providecommand{\Description}[2][]{}
\makeatletter
\def\footnotesize{\@setsize\footnotesize{9.5pt}\viiipt\@viiipt}
\makeatother
\makeatletter
\g@addto@macro\@floatboxreset{\centering}
\makeatother

\newcommand{\sys}{\textsc{Heirloom}\xspace}      %

\definecolor{oldfill}{HTML}{F0D9C4}
\definecolor{newfill}{HTML}{C9D6EA}
\newif\ifpn \pntrue

\newcommand{\na}{\textemdash}
\newcommand{\nCPUPathSlowdown}{4}
\newcommand{\nCostBatteryMAh}{6,530}
\newcommand{\nCostBatteryVsChargerPctMax}{19}
\newcommand{\nCostBatteryVsChargerPctMin}{5}
\newcommand{\nCostBridgeMax}{1.05}
\newcommand{\nCostBridgeMin}{0.86}
\newcommand{\nCostEndLevelDrop}{18}
\newcommand{\nCostEndMin}{38}
\newcommand{\nCostEndPct}{20}
\newcommand{\nCostEndPctFrom}{80}
\newcommand{\nCostEndVolt}{4.11}
\newcommand{\nCostEndkJ}{19.4}
\newcommand{\nCostEndmAh}{1,312}
\newcommand{\nCostFuelGaugeUpdates}{117}
\newcommand{\nCostGDNFusedTwoKDTokPerSec}{88}
\newcommand{\nCostGDNUnfusedTwoKDTokPerSec}{56}
\newcommand{\nCostJPerTokenTwoKD}{0.086}
\newcommand{\nCostLoadWMax}{8}
\newcommand{\nCostLoadWMin}{7}
\newcommand{\nCostNoFanCPUMax}{85.3}
\newcommand{\nCostNoFanFanCPUMax}{69.9}
\newcommand{\nCostNoFanFanNPUMax}{58.1}
\newcommand{\nCostNoFanFanSec}{2,175}
\newcommand{\nCostNoFanNPUMax}{77.5}
\newcommand{\nCostNoFanSec}{2,371}
\newcommand{\nCostNoFanSlowerPct}{9}
\newcommand{\nCostOverheadPct}{10.3}
\newcommand{\nCostPublicSelFourEpochs}{16}
\newcommand{\nCostPublicSelTwoEpochs}{4}
\newcommand{\nCostPublicSelTwoHours}{4.5}
\newcommand{\nCostRateModelFixedPct}{18}
\newcommand{\nCostRateModelRetrainPct}{4.5}
\newcommand{\nCostRateModelSelPct}{33}
\newcommand{\nCostRetrainEndPct}{5.3}
\newcommand{\nCostRetrainEndVolt}{3.95}
\newcommand{\nCostRetrainEndkJ}{4.9}
\newcommand{\nCostRetrainEndmAh}{344}
\newcommand{\nCostSelExtraPctFour}{4.8}
\newcommand{\nCostSelExtraPctTwo}{3.7}
\newcommand{\nCostThirteenBatteryMAh}{4,959}
\newcommand{\nCostThirteenJPerTokenKD}{0.173}
\newcommand{\nCostThirteenRateModelRetrainPct}{26}
\newcommand{\nCostThirteenTwoRetrainMin}{21}
\newcommand{\nCostThirteenVolt}{3.63}
\newcommand{\nCostVoltFour}{3.92}
\newcommand{\nCostVoltTwo}{4.10}
\newcommand{\nEquivAdapters}{28}
\newcommand{\nEquivBothPhonesAdapters}{4}
\newcommand{\nEquivBothPhonesUsers}{2}
\newcommand{\nEquivFixedAdapters}{27}
\newcommand{\nEquivFixedExact}{12}
\newcommand{\nEquivIdentical}{16}
\newcommand{\nEquivItemsSame}{891}
\newcommand{\nEquivItemsTotal}{919}
\newcommand{\nEquivRerunGroups}{4}
\newcommand{\nEquivRerunRuns}{9}
\newcommand{\nEquivSelOtherAdapters}{2}
\newcommand{\nEquivSelRuns}{8}
\newcommand{\nEquivSelSameLength}{6}
\newcommand{\nEquivTeacherArgmaxPublic}{485}
\newcommand{\nEquivTeacherArgmaxPublicUsers}{4}
\newcommand{\nEquivTeacherArgmaxStoredSame}{2,234}
\newcommand{\nEquivTeacherArgmaxStoredTotal}{2,260}
\newcommand{\nEquivTeacherArgmaxSynth}{490}
\newcommand{\nEquivTeacherFourUsers}{5}
\newcommand{\nEquivTeacherKL}{0.0022}
\newcommand{\nEquivTeacherTwoUsers}{19}
\newcommand{\nEquivTeacherUsers}{24}
\newcommand{\nEquivWorstDiff}{5}
\newcommand{\nEquivWorstEpochs}{8}
\newcommand{\nEquivWorstGpuRerunDiff}{3}
\newcommand{\nEquivWorstItems}{23}
\newcommand{\nEquivWorstMcNemar}{0.63}
\newcommand{\nEquivWorstNet}{2}
\newcommand{\nEquivWorstPrompts}{69}
\newcommand{\nStoredMax}{128}
\newcommand{\nStoredMedian}{83}
\newcommand{\nStoredMin}{69}
\newcommand{\nStoredWholeHistUsers}{16}
\newcommand{\nbaseRatingSmallFromVendorUp}{17}
\newcommand{\nbaseRatingSmallFromVendorVsVendor}{+6.9}
\newcommand{\nbaseRatingSmallPeople}{20}
\newcommand{\nbaseTaggingSmallPeople}{20}
\newcommand{\nfourDiscAucTest}{0.995}
\newcommand{\nfourDiscAucTrain}{0.999}
\newcommand{\nfourGainPassShare}{70.0}
\newcommand{\nfourGainShort}{0}
\newcommand{\nfourRerun}{2.99}
\newcommand{\nfourSelfRerunVsMain}{+0.1}
\newcommand{\nfourSelfRerunVsMainCI}{[\textminus 0.7, +0.9]}
\newcommand{\nfourSensUsers}{19}
\newcommand{\nfourUserZeroShift}{\textminus 0.1}
\newcommand{\ngenRecipe}{1 epoch at $10^{-4}$}
\newcommand{\ngenTrainMax}{1000}
\newcommand{\ngenTrainMin}{510}
\newcommand{\nlampthreeCap}{600}
\newcommand{\nlampthreeExcludeMajority}{90}
\newcommand{\nlampthreeExcluded}{9}
\newcommand{\nlampthreeExcludedMinShare}{92}
\newcommand{\nlampthreeMaxChars}{1,500}
\newcommand{\nlampthreeMin}{60}
\newcommand{\nlampthreePeople}{22,500}
\newcommand{\nlampthreePeopleAtLeastMin}{22,375}
\newcommand{\nlampthreeTop}{100}
\newcommand{\nlampthreeTopMax}{1,010}
\newcommand{\nlampthreeTopMin}{850}
\newcommand{\nlamptwoEvalMinFilms}{92}
\newcommand{\nlamptwoPeople}{50}
\newcommand{\nlamptwoPeopleEval}{20}
\newcommand{\nlamptwoPeopleRest}{30}
\newcommand{\nlamptwoPoolFilms}{1,013}
\newcommand{\nlamptwoTop}{20}
\newcommand{\nlamptwoUsersMax}{251}
\newcommand{\nlamptwoUsersMin}{92}
\newcommand{\nlthreeFullUsers}{100}
\newcommand{\nlthreePhiRetrainItemsOneTwoEightRecipe}{16 epochs at $5{\times}10^{-5}$}
\newcommand{\nlthreePhiUsers}{20}
\newcommand{\nlthreeRetrainItemsOneTwoEightRecipe}{2 epochs at $2{\times}10^{-5}$}
\newcommand{\nlthreeUsers}{20}
\newcommand{\nphiDiscAucTest}{0.998}
\newcommand{\nphiDiscAucTrain}{0.999}
\newcommand{\nphiGainPassShare}{51.1}
\newcommand{\nphiGainShort}{5}
\newcommand{\nphiLoraHundredAltAcc}{46.0}
\newcommand{\nphiLoraHundredFreshAltAcc}{43.2}
\newcommand{\nphiLsAttnOnlyAcc}{9.2}
\newcommand{\nphiLsFolds}{4}
\newcommand{\nphiLsInvSelfChanged}{1}
\newcommand{\nphiLsInvSelfRelErrMax}{$1.8{\times}10^{-6}$}
\newcommand{\nphiLsLftAltAcc}{23.8}
\newcommand{\nphiLsMapFoldZeroAcc}{9.0}
\newcommand{\nphiLsMapVsZeroshot}{\textminus 17.4}
\newcommand{\nphiLsMlpOnlyAcc}{24.9}
\newcommand{\nphiLsNormFirst}{610}
\newcommand{\nphiLsNormRestMax}{28.3}
\newcommand{\nphiLsNormRestMin}{1.5}
\newcommand{\nphiLsRate}{$10^{-4}$}
\newcommand{\nphiLsRateOther}{$10^{-3}$}
\newcommand{\nphiLsSelfAttnOnlyVsOrig}{\textminus 25.8}
\newcommand{\nphiLsSelfMapLoss}{25.8}
\newcommand{\nphiLsSelfMapVsOrig}{\textminus 25.8}
\newcommand{\nphiLsSelfMlpOnlyVsOrig}{\textminus 11.3}
\newcommand{\nphiLsSelfNormQkvMax}{61.0}
\newcommand{\nphiLsSelfNormQkvMin}{5.1}
\newcommand{\nphiLsShortFolds}{44}
\newcommand{\nphiLsStepsMax}{21}
\newcommand{\nphiLsStepsMin}{15}
\newcommand{\nphiLsSvdErrMax}{$1.3{\times}10^{-5}$}
\newcommand{\nphiLsUserFolds}{76}
\newcommand{\nphiLsUsers}{19}
\newcommand{\nphiSelfRerunVsMain}{0.0}
\newcommand{\nphiSelfRerunVsMainCI}{[0.0, 0.0]}
\newcommand{\nphiUserZeroShift}{0.0}
\newcommand{\nphiZeroshotFoldZeroAcc}{24.7}
\newcommand{\nrankSixteenUsers}{19}
\newcommand{\nsameCopyAbs}{7.2}
\newcommand{\nsameCopyFlip}{23}
\newcommand{\nsameCopyNeg}{18}
\newcommand{\nsameRecipeLR}{$10^{-4}$}
\newcommand{\nsameUsers}{19}
\newcommand{\ntlPoolN}{2,048}
\newcommand{\ntlPoolRequests}{2,048}
\newcommand{\ntlUserFolds}{76}
\newcommand{\ntwoDiscAucTest}{0.912}
\newcommand{\ntwoDiscAucTrain}{0.910}
\newcommand{\ntwoGainPassShare}{53.1}
\newcommand{\ntwoGainShort}{0}
\newcommand{\ntwoGenItems}{2,328}
\newcommand{\ntwoGenUsers}{10}
\newcommand{\ntwoItems}{2,555}
\newcommand{\ntwoPublicFourMinusSel}{\textminus 0.3}
\newcommand{\ntwoRerun}{2.35}
\newcommand{\ntwoSelfFourMinusSel}{0.0}
\newcommand{\ntwoSelfRerunVsMain}{\textminus 0.2}
\newcommand{\ntwoSelfRerunVsMainCI}{[\textminus 1.0, +0.7]}
\newcommand{\ntwoUserZeroShift}{\textminus 0.1}
\newcommand{\ntwoUsers}{19}

\pnfalse
\labpreprintbrand{}{SMU-FIU TrustAI4S Lab}{}
\mlsystitlerunning{Heirloom: Carrying a Phone's Personalization Across Base-Model Generations}
\makeatletter
\renewcommand{\Notice@String}{Preprint.}
\makeatother
\hypersetup{pdftitle={Heirloom: Carrying a Phone's Personalization Across Base-Model Generations},
  pdfauthor={Zhuomin Chen, Jingchao Ni, Qingkai Dong, Xu Zheng, Gabriel H\"ubner Lucchesi, Mo Sha, Wei Cheng,
  Dongsheng Luo}}

\begin{document}
\twocolumn[
\mlsystitle{Heirloom: Carrying a Phone's Personalization Across Base-Model Generations}
\begin{mlsysauthorlist}
\mlsysauthor{Zhuomin Chen}{fiu}
\mlsysauthor{Jingchao Ni}{uh}
\mlsysauthor{Qingkai Dong}{uconn}
\mlsysauthor{Xu Zheng}{fiu}
\mlsysauthor{Gabriel Hübner Lucchesi}{ind}
\mlsysauthor{Mo Sha}{fiu}
\mlsysauthor{Wei Cheng}{nec}
\mlsysauthor{Dongsheng Luo}{smu}
\end{mlsysauthorlist}
\mlsysaffiliation{fiu}{Florida International University, Miami, FL, USA}
\mlsysaffiliation{uh}{University of Houston, Houston, TX, USA}
\mlsysaffiliation{uconn}{University of Connecticut, Storrs, CT, USA}
\mlsysaffiliation{ind}{Independent Researcher, USA}
\mlsysaffiliation{nec}{NEC Laboratories America, Inc., Princeton, NJ, USA}
\mlsysaffiliation{smu}{Singapore Management University, Singapore}
\mlsyscorrespondingauthor{Dongsheng Luo}{dsluo@smu.edu.sg}
\vskip 0.3in
\begin{abstract}

On-device language models can learn a user's preferences through adapters trained on private data. 
However, a base-model upgrade may change the layers or vocabulary, leaving the personal adapter incompatible.
Replacing it with a non-personal adapter loses much of the personalization, while retraining requires retaining the user's labeled history. We present \sys, a system that transfers a personal adapter to a new base model on the phone without retaining labels. \sys uses the old personal model as a teacher, collecting its predictions on stored unlabeled inputs, vendor-provided prompts, or prompts generated by the new base. It then unloads the old base and trains an adapter for the new model to match those predictions, keeping only one base model in memory at a time. 
We evaluate \sys across three generation upgrades from two model families, an architecture-preserving refresh, and three personalization tasks.
On movie tagging, migration from stored inputs maintains pre-upgrade accuracy and differs from labeled retraining by at most 1.7 percentage points across the three upgrades. On a phone's neural processing unit (NPU), a fixed-length migration for the 4B upgrade takes 38 minutes and consumes 20\% of the battery end to end, making the stored-input setting feasible during overnight charging.

\end{abstract}
]
\printAffiliationsAndNotice{}

\section{Introduction}
\label{sec:intro}

An on-device language model serves its user best once it has learned that user's habits, such as the tags they put
on a film, how they title their notes and which messages are urgent. 
Because the data reflecting these habits is private, personalization can run on the phone.
Its common form is a low-rank adapter~\cite{lora}, tens of
megabytes of weights trained on the user's history and applied on top of a frozen base model that the vendor ships
with the operating system (OS). Phone platforms already load developer- and vendor-trained adapters on their on-device base
models~\cite{apple-adapter,google-aicore}, and recent work trains personal adapters on the phone
itself~\cite{geyko-phone-ft,edgetune,xpert}.

\begin{figure}[t]
    \centering    
    \includegraphics[width=\linewidth]{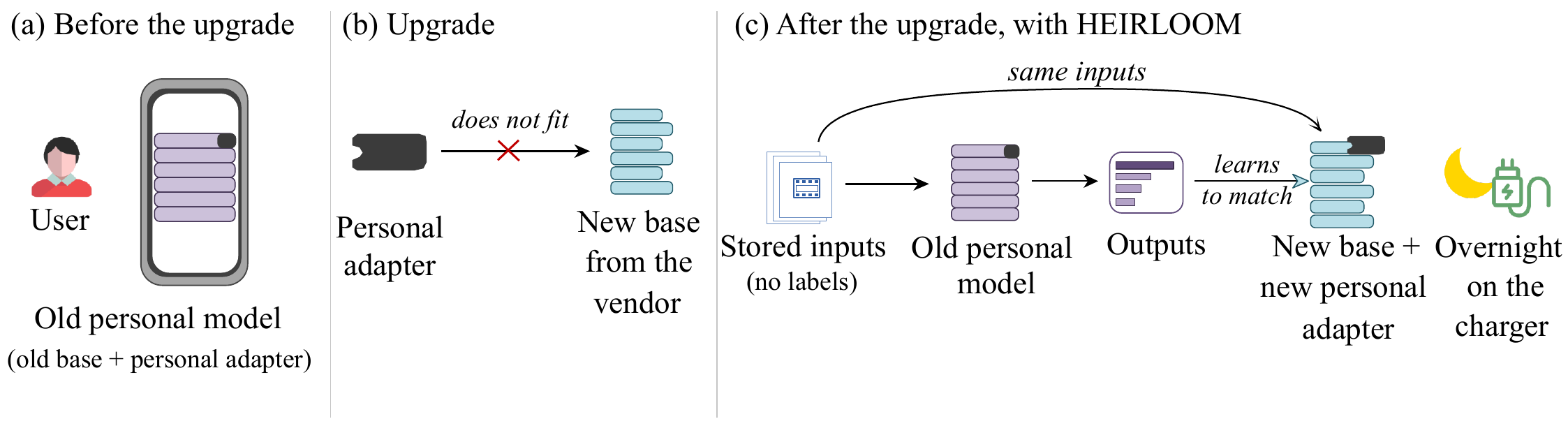}
    \caption{\sys migrates a personal adapter across a base-model upgrade. (a) The old personal model, the user's adapter
on the old base. (b) The vendor ships a new base that is incompatible with the old adapter, and retraining needs labels the phone did
not keep. (c) \sys trains the new adapter to match the old model's outputs on the stored inputs.}
    \label{fig:concept}
\end{figure}

However, the base model beneath a personal adapter does not remain fixed. Apple's on-device model changed at least three times between mid-2025 and June 2026, and both its documentation and an OS-managed foundation-model proposal describe adapters as tied to a particular base-model version~\cite{apple-adapter,apple-fm-updates,m4}. Vendors and app developers can retrain their adapters on servers when the base changes. 
Unlike vendor adapters, personal adapters cannot be retrained on a server using private user data. Retraining on the phone also becomes difficult once the original labels have been discarded.
Some upgrades retain the architecture and change only the base weights. Others introduce a new generation with different layers and tokenization. For example, Qwen3.5~\cite{qwen35} changes the number and types of Qwen3's layers~\cite{qwen3,gated-deltanet} and expands its vocabulary from about 152K to 248K tokens.
Figure~\ref{fig:concept} 
illustrates why the old adapter cannot be loaded onto the new model.

After an upgrade, a phone can fall back to the new vendor adapter, attempt to reuse the old personal adapter, or retrain it. Each option has a drawback. 
As Figure~\ref{fig:peruser} shows, the new vendor adapter lacks the user's adaptation and loses substantial accuracy in our experiments. Directly copying an adapter is not possible across an incompatible generation, and even across a compatible refresh it can change many previously correct answers. Retraining restores performance but requires access to the user's labeled history at the time of every upgrade. The inputs themselves may remain available as app content, such as films or articles, while their associated choices or edits need not be retained indefinitely. 
Keeping those labels for future retraining works against data minimization.
Moreover, preserving average accuracy is not the whole problem: an upgrade may change answers that the previous personal model got right~\cite{muscle}.

Existing adapter-transfer methods offer two broad approaches. Weight-based methods copy or transform adapter parameters between models~\cite{portllm,finetuning-transfer,paramdelta,lorasuite,cross-lora,lora-x,prolora}, but depend on some correspondence between their layers or modules. We evaluate one published mapping for the Phi upgrade~\cite{lorasuite}. In our implementation, it performs poorly even before further fine-tuning. Instead, distillation-based methods~\cite{trans-lora,titok,tuneshift-kd,upgradebench} train a new model from an old model's outputs on retained or synthetic task inputs. 
FaLA~\cite{fala} considers replacing an on-device personalized classifier, but uses labels to select training pairs and studies a smaller model.
An upgrade may invalidate a personal adapter's weights even though the old personalized model still contains the user's learned behavior. The challenge is to transfer that behavior across base-model generations after the original labels have been discarded, within the memory and energy constraints of a phone.

We present \sys, an on-device system that transfers learned user preferences across base-model generations by distilling the old personal model's behavior into a new adapter. 
Its predictions on the inputs used to train the personal adapter often reproduce the user's original labels, together with information about the teacher's confidence.
A new adapter can learn from these predictions without accessing the labels themselves. When past inputs are unavailable, the teacher can instead label task prompts that the vendor provides or the new base generates. These predictions are less directly tied to the user's past choices, but still reflect the behavior of the personalized teacher. In all cases, the new adapter begins from the vendor's adapter for the new base.

Three constraints shape the design. First, different base-model generations may use different vocabularies. Therefore, \sys transfers probabilities over task labels, or generated text for open-ended tasks. Second, the old and new bases may not fit together in phone memory. \sys records the teacher's outputs, unloads the old model, and trains the student afterward. Third, without retained labels, the phone cannot use ordinary labeled validation to decide when to stop training. \sys selects a training length using held-out agreement with the teacher. For movie tagging with stored inputs, a fixed training length performs comparably and avoids the additional selection run.

We evaluate \sys on three tasks from LaMP~\cite{lamp}: movie tagging for 19 users, product rating for 100, and headline generation for ten. We study Qwen3-1.7B to Qwen3.5-2B, Qwen3-4B to Qwen3.5-4B, and Phi-3.5-mini to Phi-4-mini, as well as an architecture-preserving refresh. Accuracy is evaluated on graphics processing units (GPUs), while two phones provide measurements of migration time, energy, and memory. On movie tagging, \sys with stored unlabeled inputs maintains the pre-upgrade accuracy across all three generation upgrades and the refresh. Its mean accuracy is within 1.7 points of fixed-recipe retraining on the three generation upgrades, and it has fewer negative flips on the 2B and Phi upgrades. 
A non-personal teacher on the same inputs performs worse, indicating that the old personal model provides the useful supervision.
When no inputs are retained, public and self-generated prompts recover much of the original performance but generally remain below stored inputs. 
On product rating, the 128-input cap reduces accuracy. Using the full unlabeled history brings \sys closer to retraining.
Although labeled retraining remains ahead on headline generation, 512 stored inputs largely preserve the old model’s ROUGE-L.

The device measurements show both the feasibility and the cost of migration. On the OnePlus 15, a stored-input migration for the 4B upgrade with a fixed training length takes 38 minutes and consumes 20\% of the battery end to end. Training one prompt for one epoch costs about the same as labeled retraining, but \sys uses more epochs and takes longer overall. Selecting a training length adds another training run. With no retained history, the larger public or self-generated prompt sets further increase the time and energy required. All measured stored-input tagging migrations fit within an eight-hour charging window on the two phones, whereas some configurations with no retained inputs do not.

We summarize our contributions as follows:
\begin{itemize}
  \item We characterize how base-model generation changes and an architecture-preserving refresh disrupt personal adapters, measuring both per-user accuracy loss and changes to previously correct predictions.
  \item We design \sys, an on-device migration system that transfers task-level teacher supervision across model vocabularies, trains with one base model in memory at a time, and selects training length without retained user labels.
  \item We evaluate \sys across three generation upgrades from two model families, an architecture-preserving refresh, and three personalization tasks. We also analyze retention budgets, alternative prompt sources, the contribution of the personal teacher, and migration time, energy, and memory on two smartphones.
\end{itemize}

\section{Background and Motivation}
\label{sec:motivation}

In this section, we first describe how personal adapters are trained and why model upgrades make them difficult to reuse. Then we compare the available fallback options and motivate learning from the old personal model without retaining labels.

\subsection{Personal adapters on the phone}
\label{sec:bg-adapters}

Low-Rank Adaptation (LoRA)~\cite{lora} leaves the base model's weights frozen. To each chosen linear layer
$W\in\mathbb{R}^{d_\mathrm{out}\times d_\mathrm{in}}$ it adds a trainable update $BA$ with
$B\in\mathbb{R}^{d_\mathrm{out}\times r}$, $A\in\mathbb{R}^{r\times d_\mathrm{in}}$ and a small rank $r$. With $r=16$ on
every projection of Qwen3-1.7B, the adapter holds 17.4M parameters, 35\,MB in 16-bit precision, against 1.7B in the
base. 
A single base can serve many adapters, each trained where the data resides.

We consider personalization in two stages. The vendor first trains a task adapter on publicly available labeled examples, such as movie tags, and distributes it with the base model. We refer to this as the vendor's adapter. The phone then fine-tunes that adapter on the user's labeled history. The resulting personal adapter combines task knowledge from public data with the user's preferences.

\subsection{Why upgrades disrupt adapters}
\label{sec:bg-upgrade}

\begin{table}[t]
\caption{Base models of the three upgrades and the refresh, from each model's configuration. Linear-attention layers are
Gated DeltaNet layers~\cite{gated-deltanet}.}
\label{tab:models}
\centering
\footnotesize
\setlength{\tabcolsep}{4pt}
\begin{tabular}{@{}lccc@{}}
\toprule
Model & Layers & Attn.\,/\,linear & Vocabulary \\
\midrule
Qwen3-1.7B & 28 & 28 / 0 & 151,936 \\
Qwen3.5-2B & 24 & 6 / 18 & 248,320 \\
\midrule
Qwen3-4B & 36 & 36 / 0 & 151,936 \\
Qwen3.5-4B & 32 & 8 / 24 & 248,320 \\
\midrule
Phi-3.5-mini & 32 & 32 / 0 & 32,064 \\
Phi-4-mini & 32 & 32 / 0 & 200,064 \\
\midrule
Qwen3-4B-Instruct-2507 & 36 & 36 / 0 & 151,936 \\
\bottomrule
\end{tabular}
\end{table}

\begin{figure}[t]
    \centering    
    \includegraphics[width=\linewidth]{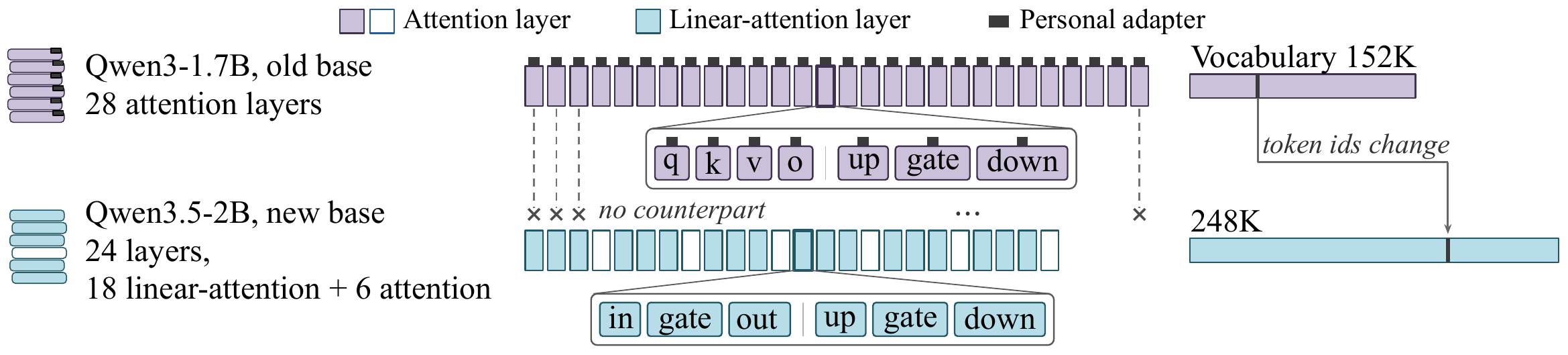}
    \caption{The step from Qwen3-1.7B to Qwen3.5-2B, drawn from the configurations in Table~\ref{tab:models}. The adapter
attaches to the projections of each of the old base's attention layers. The new base has fewer layers, three of every
four of them linear-attention layers with other projections, and a larger vocabulary whose token ids differ. The
feed-forward projections keep their shape.}
    \label{fig:generation}
\end{figure}

From Table~\ref{tab:models} and Figure~\ref{fig:generation}, we find the Qwen3-to-Qwen3.5 upgrades change several features that an adapter depends on. The number of layers decreases, and three of every four attention layers are replaced by linear-attention layers that do not have the same query, key, and value projections. Even projections that retain their dimensions are now embedded in a different network, where their activations may no longer match those seen during personalization. Weight-mapping methods that align corresponding modules~\cite{lorasuite,cross-lora,lora-x,prolora} face missing or changed counterparts. 
The vocabulary expands by 63\%, with different token IDs, making direct comparison of token-level outputs unsuitable without alignment.

The step from Phi-3.5-mini to Phi-4-mini, in a second model family, keeps the layer count and the width, grows the
vocabulary sixfold and reorganizes the attention heads. 
Only some of the old adapter’s matrices have a counterpart of the same shape.
We evaluate this partial copy in Section~\ref{sec:eval-samearch}.

A refresh such as Qwen3-4B to Qwen3-4B-Instruct-2507 keeps the architecture and the vocabulary and changes the weights
through further training. 
The old adapter still loads unchanged, and we next examine whether it continues to serve the user.

\subsection{The cost of losing personalization}
\label{sec:bg-cost}

We analyze three options available when the base changes: using the new vendor adapter, copying the old adapter where possible, and retraining from labeled history. We study personalized movie tagging in LaMP~\cite{lamp}, where the model predicts one of 15 user-specific tags from a film description. Nineteen users have between 92 and 251 films each. We train personal adapters on three quarters of each user's history and evaluate on the remaining quarter, rotating the test fold four times.

\begin{figure}[t]
\centering
\includegraphics[width=\textwidth]{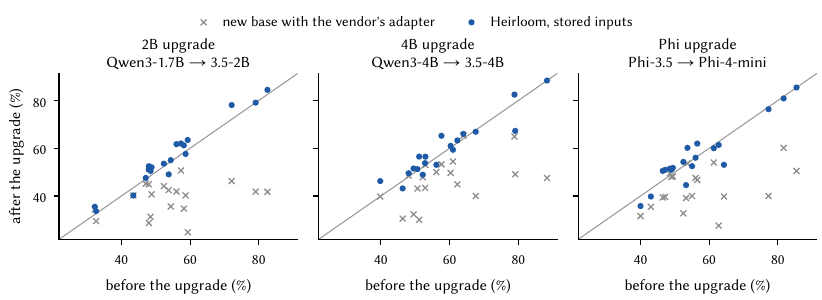}
\caption{Per-user accuracy before and after each upgrade. Crosses, the new base with the vendor's adapter, fall below
the diagonal for most users. Dots, \sys with stored inputs, lie on both sides of it. One point per user, four folds
pooled.}
\Description{Three scatter plots, one for each of the three upgrades, of each user's accuracy after the upgrade against
the same user's accuracy before it, on the same axis range, with the diagonal drawn. Crosses mark the new base with the
vendor's adapter, dots the new base after migration from the user's stored inputs.}
\label{fig:peruser}
\end{figure}

Starting over with the new vendor adapter reduces accuracy.
Relative to the old personal model, accuracy falls by 15.4 points on the 2B upgrade and 12.1 points on the 4B upgrade. 
As shown in Figure~\ref{fig:peruser}, the vendor adapter performs worse for 18 of 19 users on the 2B upgrade and 15 of 19 on the 4B upgrade.
The task adapter alone does not preserve the individual adaptation learned on the old model.

Copying fails even where the adapter loads, because the refresh changes the weights and the update $BA$ no longer
matches the inputs each layer now receives. Across the refresh, the copied adapter is \nsameCopyAbs{} points
less accurate than the old personal model, worse for \nsameCopyNeg{} of \nsameUsers{} users, and changes
\nsameCopyFlip\% of the answers the old model got right.

Retraining recovers the lost accuracy, but it requires labeled history. Fine-tuning the new vendor adapter using one learning rate and training length for all users ends 2.4 and 2.3 points above the old personal model on the two Qwen upgrades. To make this option available at each future upgrade, the phone must continue to retain the inputs together with the user's labels. Those labels record personal actions and preferences that the application may otherwise have no reason to store.

\subsection{The old model as a teacher}
\label{sec:bg-insight}

\begin{figure}[t]
\centering
\includegraphics[width=\textwidth]{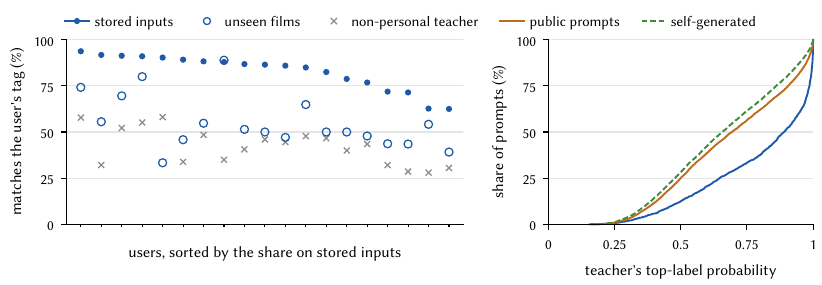}
\caption{The old personal model's outputs on past inputs largely reproduce the user's labels. Left, for each user, the
share of stored inputs whose teacher top label is the user's tag (dots), its accuracy on unseen films (circles) and the
non-personal teacher's share (crosses). Right, the teacher's top-label probability by prompt source. 2B upgrade, fold 0,
19 users.}
\label{fig:carry}
\end{figure}

At upgrade time, the old personal model is still available and can serve as a teacher. Knowledge distillation~\cite{hinton-kd} trains a student using the teacher’s predictions without requiring labeled examples.  
Figure~\ref{fig:carry} shows that, on previously seen inputs, the old personal model predicts the user’s labels more often than on unseen films and assigns higher confidence than on public or self-generated prompts.
These observations suggest that its predictions contain information useful for recovering personalization. The remaining questions are how well that information transfers to a different base, which prompts are needed, and what the migration costs on a phone.

\section{Migration Setting}
\label{sec:problem}

\noindent\textbf{Setting.} 
At upgrade time, the phone has access to the old base and its personal adapter, as well as the new base and its vendor adapter. \sys does not assume that both base models can fit in memory at the same time. 
The goal is a personal adapter for the new base whose answers serve the user at least as well as the old personal model did, while nothing about the user leaves the phone. 

\noindent\textbf{Retention levels.} 
We distinguish three levels of retained history. With inputs and labels, the phone can retrain the new adapter directly. This serves as our labeled reference. With inputs only, it retains past task inputs but not the associated user choices. For movie tagging, these are film descriptions without tags, with a default cap of 128. With no history retained, it has neither inputs nor labels, but it can still use task prompts distributed by the vendor or generated by the new base. 
Here, retained history refers only to raw task inputs and user labels. The old personal model remains available during migration at all three retention levels.

\noindent\textbf{What the levels protect.} 
These retention levels differ in the raw personal records that remain accessible to the application. 
Although keeping fewer records reduces the amount of stored history, it does not eliminate privacy risks. Membership-inference attacks can still target a LoRA-tuned model~\cite{lora-leak}, and the personal adapter retains information learned from the user’s labels.
\sys keeps the personal adapter and the teacher’s outputs on the phone and deletes the temporary outputs after migration.

\noindent\textbf{Metrics.} 
We evaluate a migration by three metrics. The first is the change in the user's task accuracy
relative to the old personal model, computed per user and averaged over users, so that a few users with long histories
do not dominate. The second metric is the \emph{negative flip rate}~\cite{pc-training,muscle}. Among the test items the
old personal model answered correctly, it is the fraction that the model after the upgrade gets wrong. The third is
what a migration costs the phone in time, energy and peak memory, measured against an overnight charge, and we compare
it with labeled retraining on the same phone.

\section{Design}
\label{sec:design}

\begin{figure}[t]
    \centering    
    \includegraphics[width=\linewidth]{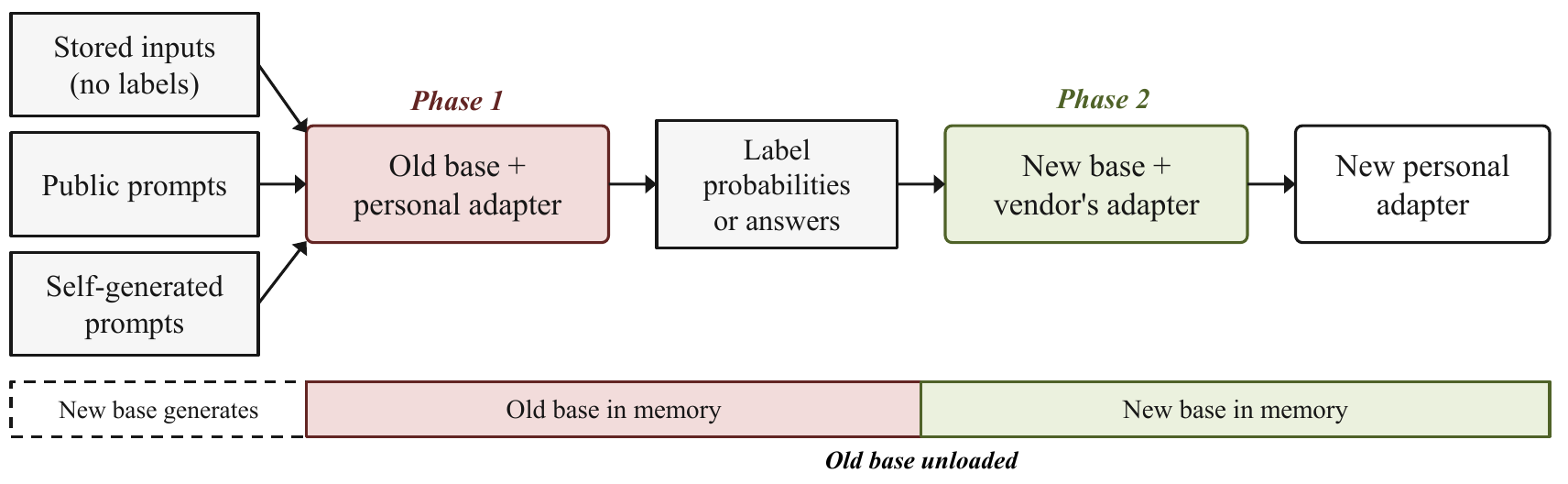}
    \caption{\sys at a base-model upgrade. Phase 1 runs the old personal model once over the prompts available at the
user's retention level and stores its outputs. Phase 2 trains a new adapter on the new base, starting from the vendor's
adapter, to reproduce those outputs. The strip below shows the one base in memory at each step.}
    \label{fig:overview}
\end{figure}

\subsection{Overview}
\label{sec:design-overview}

Figure~\ref{fig:overview} illustrates \sys's two-phase migration pipeline. The phone obtains prompts from retained inputs, vendor-provided task data, or the new model, depending on what history remains. In the first phase, the old personal model produces label probabilities or text answers for these prompts. \sys stores those outputs and unloads the old base. In the second phase, a new adapter initialized from the vendor's adapter learns to reproduce the stored outputs. When necessary, a held-out portion of the prompts determines the training length without labels. Only one base model needs to be loaded at a time. After migration, the new personal adapter replaces the old one, and the temporary teacher outputs are removed.

\subsection{Distilling label probabilities}
\label{sec:design-kd}

\textbf{Closed label sets.} When the task chooses among labels $\mathcal{L}$, such as 15 film tags, the teacher's
output for a prompt $x$ is a probability for each label, 15 numbers per film. \sys obtains them from a single forward
pass. At
the answer position it takes the model's score (logit) for each label's first token and normalizes over the labels:
\[
p_T(\ell\mid x)=\frac{\exp z_T(x)_{\mathrm{first}_T(\ell)}}{\sum_{\ell'\in\mathcal{L}}\exp z_T(x)_{\mathrm{first}_T(\ell')}}.
\]
The student computes probabilities over the same task labels using its own tokenizer and minimizes the Kullback–Leibler (KL) divergence $\mathrm{KL}(p_T\,\|\,p_S)$, averaged over prompts.
For movie tagging, the 15 labels have distinct first tokens in both tokenizers, which \sys checks once per upgrade. This allows the two models to compare label probabilities without aligning their full vocabularies. At inference, both models select the label with the highest full-sequence likelihood. Standard token-level distillation compares next-token distributions and requires vocabulary alignment, which these models do not have.
For labels sharing an initial token, we use the first position at which their token sequences differ, as detailed in Appendix~\ref{sec:app-rating}.

Matching the full label-probability distribution preserves more of the old model than matching only its top label.
Where the user's
history is ambiguous (a film the user might tag ``dark comedy'' or ``violence''), the probabilities record how the old personal model
weighs the two, and the student learns these preferences.
On prompts not used for personalization, the teacher provides its own predictions, which need not reproduce the user’s original labels.

\noindent\textbf{Generation.} When the answer is free text, such as a headline, \sys uses sequence-level
distillation~\cite{seq-kd}. The teacher writes its answer greedily, at most 32 tokens. The student is fine-tuned on it
with cross-entropy in its own tokenizer, and the generated text serves as the interface between the two vocabularies.

\subsection{Prompt sources}
\label{sec:design-prompts}

\textbf{Stored inputs.} 
When the phone retains only inputs, \sys uses them as prompts.
A cap of $B=128$ stored
inputs covers most users' whole history on movie tagging
and truncates every history on
product rating.
Headline generation, with \ngenTrainMin{} to \ngenTrainMax{} articles per
user, uses a cap of $B=512$.

\noindent\textbf{Public prompts.} With nothing kept, \sys uses $B=512$ public prompts, drawn from the data the vendor's adapter
was trained on and deduplicated. 
The public prompts contain no user-specific information, but the teacher’s predictions may reflect the user’s preferences. 

\noindent\textbf{Self-generated prompts.} If the vendor ships no prompts, the new base generates them itself, and we call these
self-generated prompts. For the 15 tags, \sys requests 512 film descriptions from it, each fitting a given tag or tag
pair. Generating them decodes up to 120 tokens per prompt, without the vendor's adapter, before phase~1.

\subsection{Selecting training length without labels}
\label{sec:design-stop}

Without labels, training length cannot be selected using ordinary validation accuracy. Agreement on training prompts is also unreliable because it tends to improve throughout training. \sys instead holds out one fifth of the prompts and evaluates checkpoints after 2, 4, 8, and 16 epochs. It selects the number of epochs with the lowest held-out KL divergence, then trains a new adapter on all prompts for that length. We call the preliminary run the selection run.
For headline generation, \sys selects the training length using the negative log-likelihood of the teacher's answers on held-out prompts, with one epoch included among the candidates.

For stored-input movie tagging, Section~\ref{sec:eval-stop} shows that a fixed eight-epoch run performs comparably to the selected length, allowing the phone to omit selection.
Selection remains useful when longer training reduces accuracy, as with public prompts for headline generation.

\subsection{Two-phase execution}
\label{sec:design-twophase}

To fit migration within phone memory, \sys separates teacher inference from student training. The old personal model first produces compact task-level targets and is unloaded before the new base is loaded. For 128 movie-tagging prompts, these targets occupy only about 3.8 KB and can also be reused for training-length selection. This allows migration to proceed with one base model in memory at a time.

For $B$ prompts and $E$ final training epochs, phase 1 requires $B$ teacher forward passes and phase 2 processes $B\cdot E$ training examples. Selection adds a 16-epoch run on four fifths of the prompts before the final run. Thus, the prompt cap bounds migration cost independently of the total history length. Labeled retraining processes the same number of examples per epoch but uses its own selected number of epochs: eight for the 2B upgrade and two for the 4B upgrade. When no inputs remain, \sys instead trains on 512 prompts, compared with a median of 83 stored inputs for tagging. Self-generated prompts add up to $512\cdot120$ decoded tokens and require loading the new base before the teacher phase.

\section{Implementation}
\label{sec:impl}

We implement \sys using PyTorch, Hugging Face Transformers, and Parameter-Efficient Fine-Tuning (PEFT). Model accuracy is evaluated on NVIDIA RTX PRO 6000 GPUs, while migration time, energy, and memory are measured on phones.
Two implementation details shorten the teacher pass and fit training into the phone's free memory. First, on tagging and rating the teacher pass needs only the
logits at the answer position, so it costs one forward pass over the prompt, a prefill. Second, when memory runs short
the student's batch is split into micro-batches with gradient accumulation, so the update is the same batch-mean loss
whatever memory is free.

On the phone, \sys uses the NPU training engine of \citet{npu-training} for both distillation and labeled retraining. The engine stores most frozen base weights in 8-bit form on the NPU and performs their matrix operations there, with precision exceptions described in Appendix~\ref{sec:app-energy}. LoRA parameters, the optimizer, and other operations, including the linear-attention recurrence, run on the central processing unit (CPU). Each update uses the same effective batch size of 16, learning rate, and training length as the GPU runs. We use the released Qwen3.5-2B and Qwen3.5-4B as students and the GPU-trained personal adapters as teachers. 
Measurements were taken on a OnePlus 15 and a OnePlus 13T, both with 16 GB of memory. Appendix~\ref{sec:app-energy} provides their detailed configurations.

\section{Experimental Setup}
\label{sec:eval-setup}

All three workloads come from the public LaMP benchmark~\cite{lamp} (its tasks LaMP-2M, LaMP-3 and LaMP-4), so no
participants were recruited and no new data was collected.

\subsection{Tagging workload} 
Personalized movie tagging~\cite{lamp} asks which of 15 tags a user would give a film from
its description. 
LaMP contains multiple question–history entries that may correspond to the same person. Following the overlap in these histories, we merge entries that share more than half their historical items and treat each merged group as one user. This differs from prior work that treats each entry as a separate user~\cite{oppu}.
After merging, the task contains \nlamptwoPeople{} users, \nlamptwoPeopleEval{} with at
least \nlamptwoEvalMinFilms{} films. The other \nlamptwoPeopleRest{} users contribute the \nlamptwoPoolFilms{} films of
the public data, and we evaluate the \nlamptwoTop{} users with the most films. 
The largest user, with 3,190 films, 37\% under one tag, and 56\% of all test items, is excluded as an outlier. Appendix~\ref{sec:app-user0} reports the results for this user.
The remaining 19 users have
\nlamptwoUsersMin{} to \nlamptwoUsersMax{} films each, \ntwoItems{} in total. 

Because the release has no usable dates, we randomly divide each user’s films into four folds, using one fold for testing and the other three for training. We rotate the test fold across four runs. The stored inputs are randomly selected from each training history, serving as a proxy for retained records.
They are at most 128, from
\nStoredMin{} to \nStoredMax{} per user and fold (median \nStoredMedian{}), the whole history for
\nStoredWholeHistUsers{} of the 19 users.

\subsection{Generation workload} 
News-headline generation~\cite{lamp} asks for the headline a user would write for an
article. We use the 10 users with the longest histories and shuffle their articles as in tagging. The last quarter
(\ntwoGenItems{} articles in all) is the test set, the rest the history, and the stored inputs are the history's last
articles. Headlines are scored by ROUGE-L F1, the longest-common-subsequence overlap with the user's own headline, on a
0 to 100 scale, as in LaMP.

\subsection{Product rating} 
We use LaMP-3~\cite{lamp}, predicting a user's rating from one to five based on a product review. 
After merging entries belonging to the same person, we select the 100 users with the most reviews, as detailed in Appendix~\ref{sec:app-rating}.
All 100 are evaluated with the main methods, while the first 20 are evaluated with every method. 
For each user, 600 randomly chosen reviews are divided into four folds, leaving 450 training reviews per fold. The 128-input cap thus truncates every history.
Public prompts come from 4,000 reviews written by other users, with no overlap in reviews. For self-generated prompts, the new base produces 512 reviews, roughly balanced across the five scores. We report accuracy, negative flip rate, and mean absolute error. 

We evaluate an upgrade only when its old base passes the personalization-usefulness rule defined in Appendix~\ref{sec:app-gate}: an adapter trained from scratch must improve mean accuracy over the vendor adapter by at least three points and help more than half the evaluated users. The 1.7B base fails the second condition for rating, so that task is evaluated on the 4B and Phi upgrades.

\subsection{Upgrades} 
We study three upgrades from two model families: Qwen3-1.7B to Qwen3.5-2B, Qwen3-4B to Qwen3.5-4B, and Phi-3.5-mini to Phi-4-mini. We call these the 2B, 4B, and Phi upgrades. We also study a refresh from Qwen3-4B to Qwen3-4B-Instruct-2507. Table~\ref{tab:models} summarizes their configurations.
Tagging runs on all four with the same 19 users and four folds.
Product rating runs on the 4B upgrade and, for the first \nlthreePhiUsers{} users, on the Phi upgrade, and generation on
the two Qwen upgrades.

\subsection{Training} 
In our experiments, the vendor adapter is a task adapter trained on public benchmark examples to represent the adapter a vendor could distribute.
All adapters use LoRA rank 16 with alpha 32. They are attached to the attention and multilayer perceptron (MLP) projections and, for Qwen3.5, to the input, gate, and output projections of the linear-attention layers. Model weights use bfloat16, and label scores use float32. We train the vendor adapter for each base and task on public data (1,013 films, 4,000 reviews, or 4,000 articles) for two epochs at a learning rate of $2\times10^{-4}$ with batches of 16. The old personal adapter is then fine-tuned on the user's history using the same recipe. Retraining on the new base uses a recipe selected by validation, whereas \sys retains the original learning rate and selects only the number of epochs, unless otherwise stated. 

Self-generated movie prompts are sampled at temperature 1 and deduplicated, with 17 descriptions per tag and 257 covering the 105 tag pairs. Some films recur across LaMP users, so public training data may contain the same film with another user's tag. This exposure is shared by all methods through their vendor adapters. The public prompts used for distillation exclude the evaluated user's items.

\subsection{Baselines} 
Except for Copy, all methods on the new base start from the vendor adapter. The vendor-adapter baseline uses that adapter without personalization. Copy loads the old personal adapter on an architecture-compatible refresh. Retrieval follows LaMP's BM25-based approach, placing the four or eight most similar labeled history examples in the prompt. 
 
Labeled retraining fine-tunes the new vendor adapter on the user's history. 
Appendix~\ref{sec:app-p0grid} reports the learning-rate and epoch grid used to select the fixed retraining recipe based on pooled first-fold validation accuracy, with ties favoring fewer epochs and then the lower learning rate.
A separate per-user tuning baseline selects the recipe for each user and fold based on that user's other three folds. Finally, the non-personal-teacher control performs the same stored-input distillation as \sys but uses the old base with its vendor adapter as the teacher. This isolates the contribution of the personal teacher from that of the prompts and distillation procedure.

\subsection{Run-to-run noise} 
GPU training is not bit-reproducible, so the same retraining configuration run twice on the
same user and fold can give two different test accuracies. Over the configurations we ran twice, the two runs differ by
\ntwoRerun{} and \nfourRerun{} points on average on tagging with the two Qwen upgrades. 
To account for this variability, we average performance across users and folds and pair tests by user.
The vendor-adapter sensitivity analysis in Appendix~\ref{sec:app-prior} and the adapter-rank analysis in Appendix~\ref{sec:app-rank} use only the first fold.
The generation workload uses one split of \ntwoGenUsers{} users.

\subsection{Metrics and comparisons} 
We compute accuracy over the test items of each user's folds and then average across users. We report the mean per-user difference $\Delta$ relative to the old personal model and a 95\% bootstrap confidence interval obtained by resampling users. Negative flip rate is the fraction of previously correct test predictions that become incorrect after an upgrade, averaged over users. We also report the cost of migration on the phone. 

For our operational comparisons, we classify a method as maintaining pre-upgrade accuracy if its mean $\Delta$ is at least $-2$ points, its confidence-interval lower bound is at least $-4$, and the sign test does not detect a loss at $p<0.05$. We describe two methods as on par under the corresponding mean-and-interval criterion, or when a Wilcoxon test detects a gain at $p<0.05$. These are practical decision criteria, not formal equivalence or non-inferiority tests. The criteria and three primary comparisons per upgrade were specified before computing the four-fold results. The two-point threshold is approximately one sixth to one eighth of the loss incurred by the vendor adapter on tagging. The tagging comparisons assess stored-input \sys against the old model and both retraining baselines. The rating comparisons assess 128-input \sys against the old model and retraining, and all-input \sys against retraining. We apply Holm correction to the Wilcoxon p-values for each set of three primary comparisons and call only adjusted $p<0.05$ significant.

\section{Evaluation}
\label{sec:eval}

We evaluate whether \sys preserves user-specific behavior across model upgrades, which design choices contribute to successful migration, and what migration costs on a phone. Our study covers three generation upgrades, an architecture-preserving refresh, three personalization tasks, multiple retention budgets and prompt sources, and two smartphones. We first examine accuracy and prediction stability, then isolate the contribution of the personal teacher and key design choices, and finally assess on-device time, energy, and memory.

\subsection{Accuracy retention on movie tagging}
\label{sec:eval-main}

\begin{table}[t]
\caption{Movie-tagging accuracy and negative flip rates across three model upgrades (19 users, four folds). $\Delta$ is the mean per-user accuracy difference from the old personal model, in percentage points. Flips is the negative flip rate (\%). Dashes indicate inapplicable comparisons.}
\label{tab:main}
\footnotesize
\setlength{\tabcolsep}{4pt}
\begin{tabular}{@{}lw{r}{2.35em}w{r}{2.35em}w{r}{2.35em}w{r}{2.35em}w{r}{2.35em}w{r}{2.35em}w{r}{2.35em}w{r}{2.35em}w{r}{2.35em}@{}}
\toprule
 & \multicolumn{3}{c}{2B upgrade} & \multicolumn{3}{c}{4B upgrade} & \multicolumn{3}{c}{Phi upgrade} \\
 & \multicolumn{3}{c}{Qwen3-1.7B$\rightarrow$3.5-2B} & \multicolumn{3}{c}{Qwen3-4B$\rightarrow$3.5-4B} & \multicolumn{3}{c}{Phi-3.5$\rightarrow$Phi-4-mini} \\
\cmidrule(lr){2-4}\cmidrule(lr){5-7}\cmidrule(l){8-10}
Method & Acc. & $\Delta$ & Flips & Acc. & $\Delta$ & Flips & Acc. & $\Delta$ & Flips \\
\midrule
Before the upgrade & 54.3 &  &  & 58.9 &  &  & 57.0 &  &  \\
\midrule
\multicolumn{10}{@{}l}{\emph{No personalization}} \\
No adapter & 19.9 & \textminus 34.5 & 75 & 34.3 & \textminus 24.5 & 53 & 26.3 & \textminus 30.8 & 62 \\
Vendor's adapter & 38.9 & \textminus 15.4 & 47 & 46.7 & \textminus 12.1 & 38 & 43.3 & \textminus 13.7 & 38 \\
\midrule
\multicolumn{10}{@{}l}{\emph{Nothing kept}} \\
LoRASuite mapping~\cite{lorasuite} & \na{} & \na{} & \na{} & \na{} & \na{} & \na{} & 8.9 & \textminus 48.1 & 90 \\
Trans-LoRA filter~\cite{trans-lora} & 53.1 & \textminus 1.3 & 15 & 56.8 & \textminus 2.1 & 16 & 55.5 & \textminus 1.5 & 13 \\
TuneShift-KD filter~\cite{tuneshift-kd} & 49.8 & \textminus 4.6 & 21 & 53.9 & \textminus 4.9 & 21 & 53.7 & \textminus 3.3 & 16 \\
\midrule
\multicolumn{10}{@{}l}{\emph{Inputs and labels kept}} \\
Retrieval~\cite{lamp}, 4 labeled examples & 38.7 & \textminus 15.7 & 47 & 49.8 & \textminus 9.1 & 33 & 48.0 & \textminus 9.0 & 30 \\
Retrieval, 8 labeled examples & 39.4 & \textminus 15.0 & 46 & 49.8 & \textminus 9.1 & 32 & 50.9 & \textminus 6.2 & 27 \\
LoRASuite, fine-tuned on 100 labels & \na{} & \na{} & \na{} & \na{} & \na{} & \na{} & 26.2 & \textminus 30.8 & 66 \\
Retraining, fixed recipe & 56.7 & +2.4 & 21 & 61.2 & +2.3 & 15 & 57.2 & +0.1 & 17 \\
Retraining, per-user tuning & 57.0 & +2.7 & 19 & 59.0 & +0.1 & 18 & 57.7 & +0.6 & 16 \\
\midrule
\multicolumn{10}{@{}l}{\emph{Inputs kept}} \\
Top-label relabeling~\cite{upgradebench} & 53.3 & \textminus 1.1 & 21 & 56.2 & \textminus 2.7 & 22 & 55.0 & \textminus 2.0 & 18 \\
\rowcolor{black!6} \sys, stored inputs & 56.2 & +1.9 & 14 & 59.5 & +0.6 & 15 & 56.7 & \textminus 0.3 & 13 \\
\bottomrule
\end{tabular}

\end{table}

\begin{figure}[t]
\centering
\includegraphics[width=\textwidth]{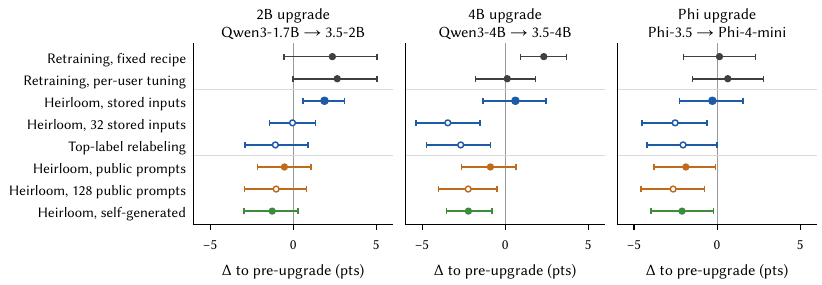}
\caption{With stored inputs and no labels, \sys keeps the pre-upgrade accuracy on all three upgrades. Dots are the
mean per-user difference to the old personal model over 19 users and four folds, bars its 95\% interval. The filter baselines reported in Tables~\ref{tab:main} and \ref{tab:filters}, together with methods outside the plotted range in Tables~\ref{tab:main} and \ref{tab:variants}, are omitted.}
\label{fig:main-ci}
\end{figure}

We first evaluate how well \sys preserves movie-tagging personalization across the three generation upgrades. Tables~\ref{tab:main} and~\ref{tab:variants} report the accuracy results, and Figure~\ref{fig:main-ci} shows their differences from the old personal model.

With stored inputs but no retained labels, \sys reaches the old personal model's mean accuracy on each of the three generation upgrades. The mean change is +1.9 points [95\% confidence interval (CI): $+0.6$, $+3.1$] on the 2B upgrade, +0.6 [$-1.3$, $+2.5$] on the 4B upgrade, and $-0.3$ [$-2.3$, $+1.6$] on the Phi upgrade. 
Although the corrected test does not establish a significant gain ($p$ = 0.06), 15 of 19 users improve and three decline on the 2B upgrade.

Across the three upgrades, \sys differs from fixed-recipe retraining by at most 1.7 percentage points in mean accuracy.
Relative to retraining with one fixed recipe on the full history, its differences are $-0.5$ [$-3.3$, +2.1], $-1.7$ [$-3.8$, +0.1], and $-0.4$ [$-1.7$, +0.7] points on the 2B, 4B, and Phi upgrades. Relative to per-user-tuned retraining, they are $-0.8$ [$-3.2$, +1.4], +0.5 [$-2.0$, +2.8], and $-0.9$ [$-3.0$, +1.0] points. 
Across all three upgrades, these comparisons satisfy our predefined practical accuracy-tolerance criterion against both retraining baselines.
This result matters when stored inputs overlap with the old adapter’s training history. On those inputs, the teacher’s output distribution retains much of the information provided by the original labels.
Appendix~\ref{sec:app-p0grid} reports the retraining configurations.

Without retained inputs, public and self-generated prompts recover much of the old model's accuracy but generally trail stored-input distillation by 1.5--3.1 points across the three upgrades. Figure~\ref{fig:carry} shows that this gap is consistent with the stronger correspondence between the teacher's predictions and the user's original labels on stored inputs. These GPU results use training lengths selected through held-out teacher agreement. For the no-history setting, four fixed epochs reduce on-device cost but lower public-prompt accuracy by a further 0.3 and 1.5 points on the two Qwen upgrades.

\subsection{The value of the personal teacher}
\label{sec:eval-teacher}

\begin{table}[t]
\caption{Movie-tagging results for alternative prompt sources and retention budgets (19 users, four folds). $\Delta$ and Flips are defined in Table~\ref{tab:main}. The shaded row repeats the stored-input result.}
\label{tab:variants}
\footnotesize
\setlength{\tabcolsep}{4pt}
\begin{tabular}{@{}lw{r}{2.35em}w{r}{2.35em}w{r}{2.35em}w{r}{2.35em}w{r}{2.35em}w{r}{2.35em}w{r}{2.35em}w{r}{2.35em}w{r}{2.35em}@{}}
\toprule
 & \multicolumn{3}{c}{2B upgrade} & \multicolumn{3}{c}{4B upgrade} & \multicolumn{3}{c}{Phi upgrade} \\
 & \multicolumn{3}{c}{Qwen3-1.7B$\rightarrow$3.5-2B} & \multicolumn{3}{c}{Qwen3-4B$\rightarrow$3.5-4B} & \multicolumn{3}{c}{Phi-3.5$\rightarrow$Phi-4-mini} \\
\cmidrule(lr){2-4}\cmidrule(lr){5-7}\cmidrule(l){8-10}
Method & Acc. & $\Delta$ & Flips & Acc. & $\Delta$ & Flips & Acc. & $\Delta$ & Flips \\
\midrule
Before the upgrade & 54.3 &  &  & 58.9 &  &  & 57.0 &  &  \\
\midrule
\multicolumn{10}{@{}l}{\emph{Nothing kept}} \\
\sys, self-generated & 53.1 & \textminus 1.3 & 16 & 56.7 & \textminus 2.2 & 15 & 54.9 & \textminus 2.1 & 15 \\
\sys, 128 public prompts & 53.3 & \textminus 1.0 & 16 & 56.6 & \textminus 2.2 & 16 & 54.4 & \textminus 2.6 & 14 \\
\sys, public prompts & 53.8 & \textminus 0.5 & 14 & 58.0 & \textminus 0.9 & 13 & 55.1 & \textminus 1.9 & 13 \\
\midrule
\multicolumn{10}{@{}l}{\emph{Inputs kept}} \\
Non-personal teacher & 43.7 & \textminus 10.6 & 37 & 43.9 & \textminus 15.0 & 40 & 43.0 & \textminus 14.1 & 37 \\
\sys, 32 stored inputs & 54.3 & 0.0 & 19 & 55.4 & \textminus 3.5 & 22 & 54.5 & \textminus 2.5 & 18 \\
\rowcolor{black!6} \sys, stored inputs & 56.2 & +1.9 & 14 & 59.5 & +0.6 & 15 & 56.7 & \textminus 0.3 & 13 \\
\bottomrule
\end{tabular}

\end{table}

The gain depends on using the personalized teacher. To isolate its contribution, we repeat stored-input distillation with the old base's non-personal vendor adapter as teacher, leaving the prompts and training unchanged. 
Table~\ref{tab:variants} shows that accuracy with the non-personal teacher falls 10.6, 15.0, and 14.1 points below the old personal model across the three upgrades.
Compared with the new vendor adapter, the non-personal teacher recovers only about a third of the lost accuracy on the 2B upgrade and none on the other two. 
This control indicates that the teacher’s learned user preferences account for most of \sys's gain.

\subsection{Preserving correct predictions}
\label{sec:eval-flips}

\begin{figure}[t]
\centering
\includegraphics[width=\textwidth]{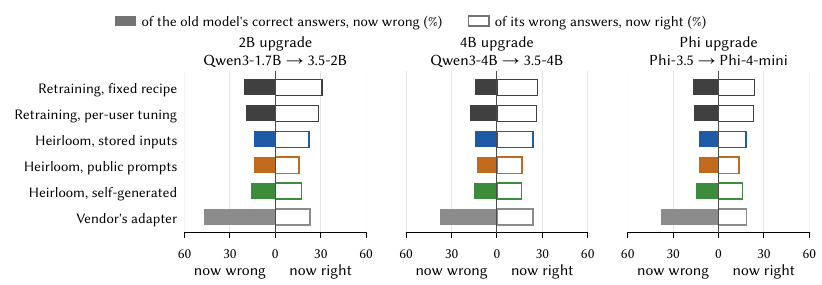}
\caption{On the 2B and Phi upgrades, \sys with stored inputs changes fewer of the answers the old model got right than either retraining baseline. Left bars give the negative flip rate of Tables~\ref{tab:main} and \ref{tab:variants}, and right bars the share of the old model's wrong answers now right, averaged over 19 users and four folds.}
\label{fig:flips}
\end{figure}

In Figure~\ref{fig:flips}, \sys preserves more of the old model's correct predictions than either retraining baseline on the 2B and Phi upgrades, while achieving a similar negative flip rate to fixed-recipe retraining on the 4B upgrade. This difference is consistent with their training objectives. Retraining learns from the user's labels, whereas \sys matches the old model's output distribution. As a result, \sys tends to retain both correct and incorrect decisions made before the upgrade, while retraining corrects more of the old model's previous mistakes.

\subsection{Prompt and training choices}
\label{sec:eval-prompts}
\label{sec:eval-stop}

\begin{figure}[t]
\centering
\includegraphics[width=\textwidth]{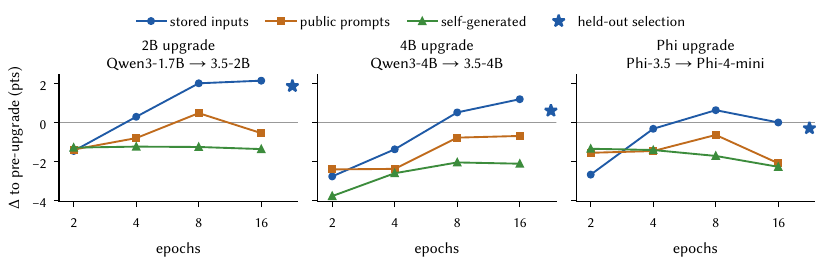}
\caption{Held-out selection (stars) lands within a point of the best single length. Lines show accuracy relative to the
old personal model after 2 to 16 epochs on all prompts, per prompt source, averaged over 19 users and four folds.}
\label{fig:epochs}
\end{figure}

\textbf{Prompt budget.} 
With only 32 stored inputs, \sys ends 0.0, -3.5, and -2.5 points from the old personal model, below the results with up to 128 inputs. Table~\ref{tab:variants} and Figure~\ref{fig:main-ci} show this prompt-budget effect.
With public prompts, 512 do better than 128 on all three upgrades, because
without the teacher's training inputs the student needs prompts that cover the task's inputs.

\textbf{Prompt selection.} Public prompts outperform the unfiltered self-generated prompts by 0.7, 1.3, and 0.2 points across the three upgrades. 
We also adapt the prompt-selection strategies of Trans-LoRA~\cite{trans-lora} and TuneShift-KD~\cite{tuneshift-kd} to choose 512 prompts from 2,048 generated candidates, while keeping \sys's distillation procedure unchanged. 
Relative to the corresponding unfiltered runs, Trans-LoRA filtering changes accuracy by +0.2 [$-0.8$, +1.1], 0.0 [$-1.0$, +1.2], and +0.6 [$-0.7$, +2.0] points, showing little difference. TuneShift-KD filtering reduces accuracy by 3.1, 2.8, and 1.2 points, respectively. 
The adapted filters provide no consistent improvement over the simpler public-prompt setting. Appendix~\ref{sec:app-filters} details their implementation and the candidate-pool differences affecting these comparisons.

\textbf{Distillation targets.} 
Table~\ref{tab:main} and Figure~\ref{fig:main-ci} show that distilling the full label-probability distribution improves accuracy by 3.0, 3.3, and 1.8 points over using only the top label, which also causes more negative flips.

\textbf{Training length.} 
As Figure~\ref{fig:epochs} shows, held-out selection approaches the best fixed training length identified retrospectively from test accuracy, with a gap of at most 0.9 points across the three upgrades. This suggests that teacher agreement can guide training length without labels. For stored-input tagging, a fixed eight-epoch run achieves nearly the same accuracy as the selected length, differing by +0.1, -0.1, and +0.9 points across the upgrades. The phone can therefore omit the additional selection run for this setting. As shown in Appendix~\ref{sec:app-gen}, selection remains useful for headline generation, where longer training can reduce accuracy.

\subsection{Limits of adapter weight reuse}
\label{sec:eval-samearch}

\begin{table}[t]
\caption{Accuracy and negative flips following the Qwen3-4B refresh (19 users, four folds). $\Delta$ is the difference from the old personal model, with 95\% confidence intervals. Retraining uses eight epochs at a learning rate of \nsameRecipeLR{}}
\label{tab:samearch}
\centering
\footnotesize
\setlength{\tabcolsep}{4pt}
\begin{tabular}{@{}lccc@{}}
\toprule
Method & Acc. & $\Delta$ [95\% CI] & Flips \\
\midrule
Before the upgrade & 58.9 & -- & -- \\
Copy the old adapter & 51.6 & \textminus 7.2 [\textminus 9.5, \textminus 5.2] & 23 \\
\sys, stored inputs & 58.1 & \textminus 0.8 [\textminus 2.2, +0.7] & 13 \\
\sys, self-generated & 55.7 & \textminus 3.2 [\textminus 5.2, \textminus 1.3] & 16 \\
Retraining, fixed recipe & 58.3 & \textminus 0.5 [\textminus 2.3, +1.1] & 17 \\
Vendor's adapter & 44.5 & \textminus 14.3 [\textminus 19.6, \textminus 9.6] & 37 \\
\bottomrule
\end{tabular}

\end{table}

Even when the architecture is unchanged, copying the old adapter can lose accuracy. 
Across the Qwen3-4B refresh, direct copying reduces accuracy by 7.2 points and changes 23\% of previously correct predictions. Direct copying reduces accuracy for 18 of the 19 users, with aggregate results reported in Table~\ref{tab:samearch}.
\sys with stored inputs instead differs from the old personal model by $-0.8$ points [$-2.2$, +0.7], with a 13\% negative flip rate. It also differs by only $-0.2$ points [$-1.7$, +1.2] from labeled retraining. Copying the old adapter as an initialization does not improve either method: relative to starting from the vendor adapter, it changes labeled retraining by $-1.0$ points and \sys by $-5.1$. The refresh changed the base weights and position encoding, altering the activations on which the old adapter had been trained. For this measured refresh, reproducing the old model's behavior is more effective than carrying its weights forward.

Partial weight reuse also performs poorly in the Phi upgrade. The attention-output and two feed-forward projections retain compatible shapes, but the fused query-key-value projection does not. Copying only the compatible components changes 107 of the 2,555 predictions made by the new base without an adapter and increases accuracy by just 0.3 points, leaving it 16.8 points below the vendor adapter. Using this partial copy to initialize labeled retraining or \sys lowers accuracy by 4.5 and 4.1 points, respectively, compared with initialization from the vendor adapter.

We evaluate an implementation of LoRASuite’s published mapping on the Phi upgrade~\cite{lorasuite}. Because its released code does not support this model pair, we implement the mapping from the paper’s equations. The mapped adapter achieves 8.9\% accuracy, compared with 26.3\% for the new base without an adapter. Fine-tuning on up to 100 labeled examples improves its accuracy to 26.2\%, still below the same fine-tuning from the vendor adapter or a fresh LoRA. Appendix~\ref{sec:app-lorasuite} examines the unusually large mapped updates underlying this result. 

\subsection{Product rating with limited history}
\label{sec:eval-rating}

\begin{table}[t]
\caption{Product rating on the 4B upgrade (Qwen3-4B $\rightarrow$ Qwen3.5-4B), four folds, all \nlthreeFullUsers{}
users and the first \nlthreeUsers{}. $\Delta$ is the per-user difference in points with its 95\% interval, against the old
personal model and against retraining on the whole labeled history. MAE is the mean absolute error of the score. Each
group's last row compares \sys with retraining at 128 inputs each.}
\label{tab:lamp3-all}
\footnotesize
\setlength{\tabcolsep}{4pt}
\begin{tabular}{@{}lrrrrr@{}}
\toprule
 & & \multicolumn{2}{c}{$\Delta$ [95\% CI] against} & & \\
\cmidrule(lr){3-4}
Method & Acc. & pre-upgrade & retraining & Flips & MAE \\
\midrule
\multicolumn{6}{@{}l}{\emph{All 100 users}} \\
Before the upgrade & 73.8 & -- & -- & -- & 0.28 \\
Retraining, fixed recipe & 75.1 & +1.3 [+0.8, +1.8] & -- & 12 & 0.26 \\
Retraining, 128 stored inputs & 73.8 & 0.0 [\textminus 0.6, +0.5] & \textminus 1.3 [\textminus 1.7, \textminus 1.0] & 14 & 0.28 \\
\sys, 128 stored inputs & 72.9 & \textminus 0.9 [\textminus 1.4, \textminus 0.5] & \textminus 2.2 [\textminus 2.8, \textminus 1.8] & 12 & 0.29 \\
\sys, all inputs & 74.5 & +0.7 [+0.3, +1.0] & \textminus 0.6 [\textminus 1.0, \textminus 0.3] & 8 & 0.27 \\
\addlinespace
Vendor's adapter & 64.0 & \textminus 9.9 [\textminus 11.8, \textminus 8.0] & \textminus 11.2 [\textminus 13.0, \textminus 9.5] & 28 & 0.40 \\
No adapter & 63.0 & \textminus 10.8 [\textminus 12.8, \textminus 9.0] & \textminus 12.1 [\textminus 14.0, \textminus 10.4] & 31 & 0.42 \\
\addlinespace
\sys{} vs.\ retraining, 128 inputs each &  &  & \textminus 0.9 [\textminus 1.4, \textminus 0.4] &  &  \\
\midrule
\multicolumn{6}{@{}l}{\emph{First 20 users}} \\
Before the upgrade & 74.9 & -- & -- & -- & 0.26 \\
Retraining, fixed recipe & 76.8 & +1.9 [+0.6, +3.3] & -- & 10 & 0.24 \\
Retraining, 128 stored inputs & 75.2 & +0.3 [\textminus 0.8, +1.5] & \textminus 1.6 [\textminus 2.2, \textminus 0.9] & 13 & 0.26 \\
\sys, 128 stored inputs & 74.8 & \textminus 0.1 [\textminus 1.0, +1.0] & \textminus 2.0 [\textminus 2.8, \textminus 1.1] & 10 & 0.26 \\
\sys, all inputs & 75.8 & +0.9 [+0.1, +1.7] & \textminus 1.0 [\textminus 2.0, 0.0] & 7 & 0.25 \\
\addlinespace
Vendor's adapter & 63.4 & \textminus 11.5 [\textminus 16.8, \textminus 6.9] & \textminus 13.4 [\textminus 18.9, \textminus 8.8] & 29 & 0.39 \\
No adapter & 62.9 & \textminus 12.0 [\textminus 17.1, \textminus 7.3] & \textminus 13.9 [\textminus 18.9, \textminus 9.4] & 31 & 0.40 \\
\addlinespace
\sys{} vs.\ retraining, 128 inputs each &  &  & \textminus 0.4 [\textminus 1.2, +0.5] &  &  \\
\bottomrule
\end{tabular}

\end{table}

We study product rating, where every user's training history is longer than the 128-input cap. 
On the 4B upgrade, we evaluate the main methods for all 100 users, as shown in Table~\ref{tab:lamp3-all}. Appendix~\ref{sec:app-rating} provides the full baseline results for the first 20 users on the 4B and Phi upgrades.
On product rating, the vendor adapter raises the new base's accuracy from 63.0\% to 64.0\%, while personalization provides a much larger gain. This highlights the value of preserving user-specific adaptation across upgrades.

\begin{figure}[t]
\centering
\includegraphics[width=\textwidth]{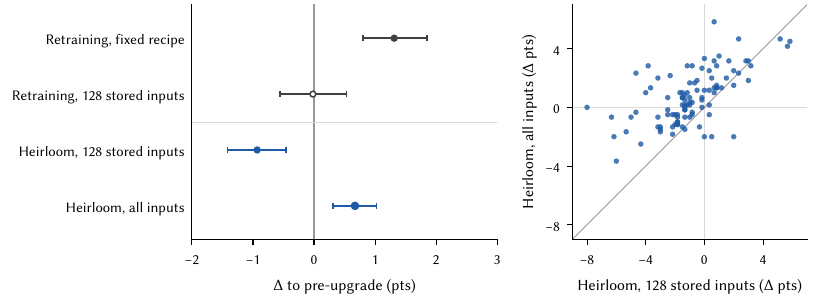}
\caption{Product rating on the 4B upgrade, \nlthreeFullUsers{} users. Left, the rows of Table~\ref{tab:lamp3-all} as
mean per-user differences to the old personal model with 95\% intervals, its vendor's adapter and no-adapter rows
below this range. Right, per user, \sys with all stored inputs against \sys at the 128-input cap, both relative to the
old personal model.}
\label{fig:rating}
\end{figure}

Table~\ref{tab:lamp3-all} shows that at the 128-input cap, \sys achieves 72.9\% accuracy, 0.9 points below the old personal model and 2.2 points below retraining on the full labeled history.
Retraining on the same 128 inputs with labels reaches 73.8\%, leaving a 0.9-point gap attributable to the different supervision under this shared input budget. When \sys has access to all 450 unlabeled training inputs, accuracy rises to 74.5\%, 0.7 points above the old personal model and 0.6 points below full-history labeled retraining. The latter difference is small under the paper's operational criterion.
The 128-input cap costs 1.6 points [1.2, 2.0] compared with all inputs. 
At that cap, 67 of 100 users fall below their pre-upgrade accuracy. Figure~\ref{fig:rating} shows that retaining all inputs improves accuracy for 86 of the 100 users relative to the capped version.
Thus, removing labels has a modest cost when the same inputs are available, while discarding most of the inputs adds a further limitation.

The effect of the cap differs between tasks because it covers different fractions of their training histories. In movie tagging, 128 inputs include the whole history for 16 of 19 users. In product rating, the same budget covers only 28\% of each user's 450 inputs, leaving the student's training set without teacher outputs for the other 72\%. This difference is consistent with the 1.6-point gain from using all inputs on rating. It also indicates why accuracy results obtained when nearly the full history is retained should not automatically be extrapolated to longer histories.
The three primary comparisons for the first 20 users are within one point of the corresponding 100-user results in Table~\ref{tab:lamp3-all}.

As reported in Appendix~\ref{sec:app-rating}, the 128-input \sys model on the Phi upgrade is 1.0 point [$-2.2$, +0.1] below the old personal model and 1.6 points [$-2.4$, $-0.8$] below full-history retraining for the first 20 rating users.
Unlike the 4B rating upgrade, the Phi vendor adapter provides a larger improvement over the new base without an adapter, with a mean gain of 7.9 points [0.0, 15.8].

\subsection{Headline generation and retraining}
\label{sec:eval-gen}

\begin{table}[t]
\caption{Headline generation on the two Qwen upgrades, 10 users. R-L is ROUGE-L F1 $\times$100 and $\Delta$ its difference
to the old personal model. Retraining uses the user's labeled history (\ngenTrainMin{} to \ngenTrainMax{} articles),
while the stored-input rows use at most 512 or 128 inputs, without labels. Best per user is each user's setting
chosen on that user's test set.}
\label{tab:lamp4}
\centering
\footnotesize
\setlength{\tabcolsep}{4pt}
\begin{tabular}{@{}llrrrr@{}}
\toprule
Personal & & \multicolumn{2}{c}{1.7B $\rightarrow$ 3.5-2B} & \multicolumn{2}{c}{4B $\rightarrow$ 3.5-4B} \\
\cmidrule(lr){3-4}\cmidrule(l){5-6}
data kept & Method & R-L & $\Delta$ & R-L & $\Delta$ \\
\midrule
-- & Before the upgrade & 12.66 & -- & 13.41 & -- \\
\midrule
inputs + labels & Retraining, fixed recipe & 13.11 & +0.45 & 14.62 & +1.21 \\
 & Retraining, best per user & 13.35 & +0.69 & 14.76 & +1.35 \\
\midrule
inputs & \sys, 512 stored & 12.31 & \textminus 0.35 & 13.63 & +0.22 \\
 & \sys, 128 stored & 11.61 & \textminus 1.05 & 13.38 & \textminus 0.03 \\
\midrule
nothing & \sys, public & 12.12 & \textminus 0.55 & 13.33 & \textminus 0.08 \\
\midrule
-- & Non-personal teacher & 10.80 & \textminus 1.86 & 12.27 & \textminus 1.14 \\
 & Vendor's adapter & 11.07 & \textminus 1.59 & 12.11 & \textminus 1.30 \\
 & No adapter & 7.66 & \textminus 5.00 & 9.00 & \textminus 4.41 \\
\bottomrule
\end{tabular}

\end{table}

\begin{figure}[t]
\centering
\includegraphics[width=\textwidth]{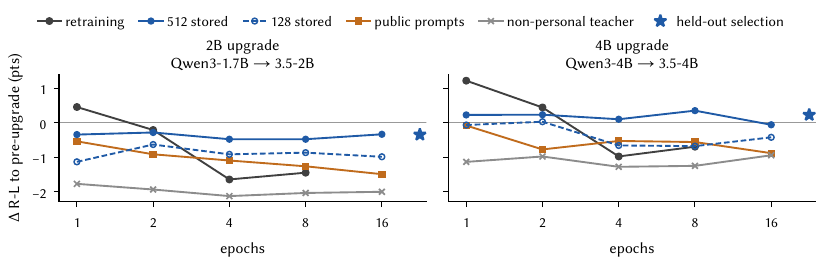}
\caption{Headline generation, ROUGE-L relative to the old personal model against training epochs on all prompts, 10
users. Lines are the cells of Table~\ref{tab:lamp4-grid}, \sys by prompt set and teacher, and retraining at its
recipe's learning rate. Stars are the held-out selection for 512 stored inputs. Each value subtracts the old personal model's
R-L of Table~\ref{tab:lamp4}.}
\label{fig:gen}
\end{figure}

As shown in Table~\ref{tab:lamp4}, \sys preserves much of the old personal model’s performance on both Qwen headline-generation upgrades by distilling its generated text. The old personal model exceeds the vendor adapter by 1.6 and 1.3 ROUGE-L points. With 512 stored inputs, \sys recovers 78\% of this personalization gain on the 2B upgrade and 117\% on the 4B upgrade, slightly exceeding the old personal model on the latter. Appendix~\ref{sec:app-gen} reports smaller gains with 128 stored inputs or public prompts. Distilling the non-personal teacher instead produces results close to the vendor adapter.

Retaining the original labels offers a further advantage on headline generation. Retraining uses each user's full labeled history of 510–1,000 articles, while \sys uses at most 512 inputs. With the fixed recipe selected across other users (\ngenRecipe{}), retraining improves on the old model by 0.45 [0.16, 0.77] and 1.21 [0.60, 1.78] ROUGE-L points. It exceeds the 512-input \sys model by 0.80 [0.43, 1.15] and 0.99 [0.56, 1.34] points on the two upgrades, respectively, with a gain for nine of the ten users in each. Figure~\ref{fig:gen} shows that retraining and public-prompt distillation can deteriorate with additional epochs, whereas stored-input distillation is comparatively stable. Thus, stored-input migration largely reproduces the old model's behavior, but labeled retraining can improve on it.

We tested whether transferring more of the teacher's output distribution narrows the headline-generation gap. 
Appendix~\ref{sec:app-uld} reports that adding universal logit distillation~\cite{uld} to the 512-input text-distillation objective changes ROUGE-L by +0.25 [$-$0.39, +0.91] and $-$0.05 [$-$0.54, +0.50] points on the two upgrades.
We observe no clear improvement in these experiments. For this open-ended task, generated teacher text preserves much of the old model's behavior, but not necessarily the additional information available from the user's original labels.

\subsection{Cost on the phone}
\label{sec:eval-cost}

\suppressfloats[t]

\begin{table}[t]
\caption{Cost of one migration on the OnePlus 15 for the user with the median number of stored inputs
(\nStoredMedian{}). Phase 1 is the old personal model's teacher pass and phase 2 the training of the new adapter on the
new base. Minutes are measured on the charger, and energy is in kJ and as a share of the battery. $\Delta$ is the accuracy at the
fixed length minus that at the selected length, from the GPU runs.}
\label{tab:cost}
\footnotesize
\setlength{\tabcolsep}{4pt}
\begin{tabular}{@{}lcccccccc@{}}
\toprule
 & \multicolumn{3}{c}{Minutes} & Tokens/s & \multicolumn{2}{c}{Energy} & Peak memory & \\
\cmidrule(lr){2-4}\cmidrule(lr){6-7}
Method & Phase 1 & Phase 2 & Total & phase 2 & kJ & \% battery & (GB) & $\Delta$ \\
\midrule
\multicolumn{9}{@{}l}{\emph{2B upgrade, Qwen3-1.7B $\rightarrow$ Qwen3.5-2B}} \\
Stored inputs, selected length & 1.0 & 71.8 & 72.7 & 77 & 29.3 & 30 & 2.6 & -- \\
Stored inputs, 8 epochs & 1.0 & 20.0 & 21.0 & 77 & 8.3 & 8.6 & 2.6 & +0.1 \\
Retraining with labels & -- & 20.0 & 20.0 & 79 & 8.2 & 8.5 & 2.5 & -- \\
\midrule
\multicolumn{9}{@{}l}{\emph{4B upgrade, Qwen3-4B $\rightarrow$ Qwen3.5-4B}} \\
Stored inputs, selected length & 1.5 & 69.9 & 71.4 & 41 & 30.7 & 33 & 3.3 & -- \\
Stored inputs, 8 epochs & 1.5 & 36.2 & 37.8 & 43 & 16.7 & 18 & 3.6 & \textminus 0.1 \\
\quad measured end to end & -- & -- & -- & -- & 19.4 & 20 & -- & -- \\
Retraining with labels & -- & 10.1 & 10.1 & 39 & 4.1 & 4.5 & 3.4 & -- \\
\quad measured end to end & -- & -- & -- & -- & 4.9 & 5.3 & -- & -- \\
Public prompts, 4 epochs & 9.1 & 116.7 & 125.8 & 42 & 54.8 & 60 & 3.6 & \textminus 1.5 \\
Self-generated prompts, 4 epochs & 10.3 & 127.3 & 137.6 & 45 & 63.8 & 69 & 3.6 & \textminus 0.4 \\
Public prompts, selected length & 9.1 & 879.5 & 888.6 & 40 & 371.9 & 404 & 4.1 & -- \\
\bottomrule
\end{tabular}

\end{table}

\begin{figure}[t]
\centering
\includegraphics[width=\textwidth]{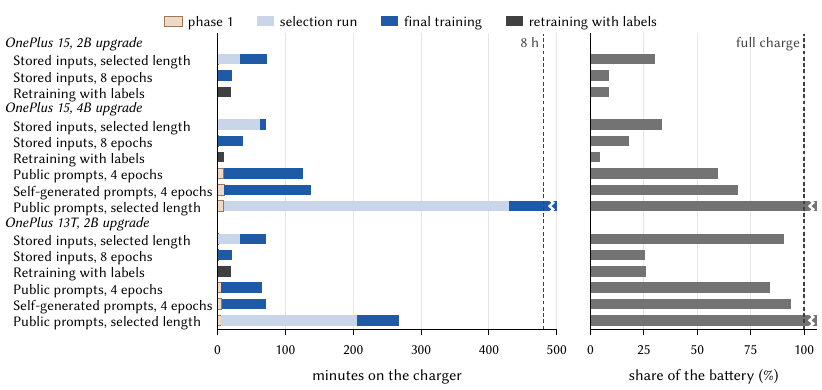}
\caption{On the two phones, every measured migration from stored inputs fits in an 8-hour overnight charge. Left,
minutes on the charger, as phase 1, selection run and final training. Right, the energy from the per-token rates, idle
draw subtracted, as a share of the battery. Bars cut at the axis end continue in Tables~\ref{tab:cost} and
\ref{tab:cost13t}.}
\label{fig:cost}
\end{figure}

We measure migration costs on the OnePlus 15 for both Qwen upgrades and on the OnePlus 13T for the 2B upgrade. Table~\ref{tab:cost} summarizes the main measurements, and Appendix~\ref{sec:app-energy} details the measurement conditions.
The timed runs use a charger, a screen-off phone, and an external cooling fan. 
Except for the end-to-end measurements, energy is estimated from per-token measurements on battery with idle consumption subtracted. Battery-share values represent charge consumption relative to the device’s full capacity, so values above 100\% indicate multiple full-charge equivalents. We also measure one 4B migration and a labeled-retraining run end to end on battery.
We measure one 4B migration and a labeled-retraining run end to end on battery. The Phi upgrade is not included in device measurements because the training engine has not been validated for its fused query-key-value projections and partial rotary position encoding.

As shown in Figure~\ref{fig:cost}, all measured stored-input tagging migrations fit within an eight-hour charging window on both phones.
A distillation epoch and a labeled-retraining epoch have similar costs: 271 versus 303 seconds on the 4B upgrade and 150 seconds for both on the 2B upgrade. Per-token energy is comparable. The overall difference comes from training length. With a fixed eight epochs, migration takes 1.05 times as long as retraining on the 2B upgrade and 3.7 times as long on the 4B upgrade. Including the selection run increases these ratios to 3.6 and 7.0. The selection run accounts for 45\% and 85\% of the measured selected-length migration time, respectively. As shown in Section~\ref{sec:eval-stop}, the fixed eight-epoch setting avoids this overhead on stored-input movie tagging with little change in accuracy.

Table~\ref{tab:cost13t} shows that on the OnePlus 13T, stored-input migration for the 2B upgrade draws 26\% of the battery at a fixed length and 91\% with selection.
Retraining takes \nCostThirteenTwoRetrainMin{} minutes and \nCostThirteenRateModelRetrainPct\% of the battery. The 13T
draws \nCostThirteenJPerTokenKD{} J per training token against \nCostJPerTokenTwoKD{} on the OnePlus 15.

\begin{figure}[t]
\centering
\includegraphics[width=\textwidth]{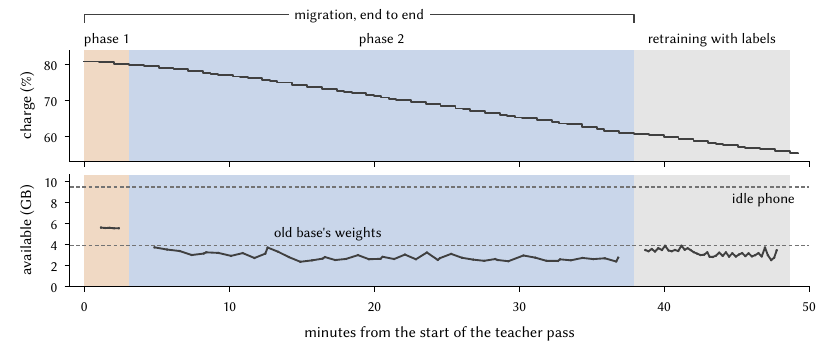}
\caption{The 4B stored-input migration on the OnePlus 15 at a fixed eight epochs, followed by labeled retraining, was measured end to end on battery. Table~\ref{tab:cost} reports the corresponding values.
Top, the charge counter as a share of the
full charge. Bottom, available memory from each job's step log, with the idle reading and the old base's weights
marked.}
\label{fig:timeline}
\end{figure}

Measured end to end on battery, the 4B upgrade's migration from stored inputs at a fixed 8 epochs takes
\nCostEndMin{} minutes and draws \nCostEndPct\% of the battery (\nCostEndkJ{}\,kJ, Figure~\ref{fig:timeline}). End to end means from the start
of the teacher pass until the next job begins, idle draw and fixed overheads included. We measured a labeled
retraining run the same way immediately afterward, and it drew \nCostRetrainEndPct\%. The fixed overheads and the
run without the fan are in Appendix~\ref{sec:app-energy}.

Keeping no task history increases migration cost.
On the 4B upgrade, 512 public or self-generated prompts require about 6.3 times as many training tokens per epoch as the median user's stored inputs. At four fixed epochs, the estimated charge consumption is 60\% for public prompts and 69\% for self-generated prompts. Their GPU accuracy is 1.5 and 0.4 points below the selected-length result, respectively. Adding held-out selection to public prompts raises the estimated running time to 14.8 hours, beyond a single eight-hour charging window. 
For the measured no-history configuration, \sys uses four fixed epochs to limit the device cost.

The 4B upgrade illustrates why two-phase execution is needed. During student training, the new base occupies 4.7 GB of NPU weights, and the phone's available memory falls from 9.4 GB at idle to 2.2 GB.
As Figure~\ref{fig:timeline} shows, the available memory is smaller than the old base’s 3.9 GB of weights, so keeping both bases resident would exceed the memory available in this configuration.
Instead, phase 1 stores only 30 bytes per movie-tagging prompt, or ten bytes per rating prompt, before unloading the teacher. On the 2B upgrade, both bases could fit under the measured conditions, but the sequential design also works there. 
Table~\ref{tab:cost} reports peak trainer-process resident memory of 2.6 and 3.6 GB for the stored-input configurations of the two upgrades, respectively.

The phone's phase~1 reproduces the GPU teacher's label probabilities to a mean KL divergence of at most
\nEquivTeacherKL{} per teacher, over the \nEquivTeacherUsers{} old personal models run on the phones. Of the \nEquivAdapters{} adapters we trained on the phones, at the selected length or with the retraining
recipe, \nEquivIdentical{} predict every test item exactly as the GPU run of the same user and fold. All of them
together agree with the GPU on \nEquivItemsSame{} of \nEquivItemsTotal{} items. 
Appendix~\ref{sec:app-equiv} reports that the two phones, with different NPUs, train bit-identical adapters in all four configurations run on both, and that reruns on one phone are also bit-identical.

\section{Discussion}
\label{sec:discussion}

\textbf{Running \sys on the phone.} 
A platform could schedule migration while the phone is charging and replace the old personal adapter after training completes. Because \sys saves the teacher’s outputs before unloading the old base, an interrupted student-training phase can restart without repeating the teacher pass. A deployment would need to coordinate migration with model installation and the phone’s normal workload.

\textbf{Implications for platforms.} 
To use \sys, a platform would need to retain the old base long enough to produce teacher outputs and make the new task adapter available before migration. 
The experiments suggest that the initialization of the vendor adapter matters. 
Appendix~\ref{sec:app-prior} shows that, in first-fold comparisons, starting from the previous recipe or an adapter tuned for public-data accuracy changes the absolute results, but does not systematically widen the gap between \sys and labeled retraining.

\textbf{Retrieval instead of an adapter.} 
A system can personalize the new model by placing labeled history examples in each prompt, avoiding adapter migration. Previous LaMP results with larger models show that retrieval can outperform per-user fine-tuning~\cite{salemi-rag-peft}. However, retrieval requires retaining the labels and increases the context processed at every request. 
In our tagging experiments, Table~\ref{tab:main} shows that using eight similar labeled films leaves the new model 6.2–15.0 points below the old personal model. Stored-input migration comes much closer without retaining labels.
Retrieval remains complementary to a migrated adapter when labeled examples are available.

\textbf{Closed label sets and open generation.} 
Task format affects the supervision available for migration. For closed-set tasks, label probabilities preserve the teacher's relative preferences among alternative answers. For open-ended generation, a generated response provides a narrower supervision target. This distinction is consistent with the headline-generation results, where \sys retains much of the old model's performance while full-history labeled retraining achieves further gains.

\textbf{Teacher quality and history coverage.} 
Migration quality depends on the personalization learned by the old model and the retained inputs available at upgrade time. On product rating, using the full unlabeled history improves accuracy by 1.6 points over the 128-input cap. This result highlights a practical trade-off between retained history, migration accuracy, and device computation.

\textbf{Evaluation scope and future deployment.} 
Our evaluation characterizes personalization retention on LaMP user histories using GPU-trained personal models, with migration costs and prediction agreement on two phones. These results establish the behavior and resource requirements of the evaluated upgrades and provide a basis for studying time-ordered histories and successive model generations.

\section{Related Work}
\label{sec:related}

\textbf{Weight-based adapter transfer.} 
Prior work reuses or transforms adaptation weights between models. PortLLM~\cite{portllm}, task-vector approaches~\cite{chat-vector,paramdelta,finetuning-transfer}, and other weight-space transfers assume conditions under which parameter updates remain meaningful across models. Cross-LoRA~\cite{cross-lora}, LoRASuite~\cite{lorasuite}, LoRA-X~\cite{lora-x}, ProLoRA~\cite{prolora} and
TransFusion~\cite{transfusion} align layers, heads, or subspaces. LoRASuite also fine-tunes the mapped adapter with task data. Other approaches, including Trans-PEFT~\cite{trans-peft}, Portable Reward Tuning~\cite{portable-reward}, and CAR-LoRA~\cite{car-lora}, change training so adapters are more portable across later model changes. Recyclable Tuning~\cite{recyclable} reuses or distills previous adaptations during continued pretraining. \sys does not depend on transferring adapter weights or preparing the original personal adapter for a future upgrade.

\noindent\textbf{Distillation-based transfer.} 
Trans-LoRA~\cite{trans-lora}, TiTok~\cite{titok} and
TuneShift-KD~\cite{tuneshift-kd} use synthetic examples and teacher predictions to transfer fine-tuned models. Their prompt generation or filtering may depend on seed examples, auxiliary models, or task-specific training. UpgradeBench~\cite{upgradebench} studies model-specialist upgrades across successive Qwen generations and finds that relabeling retained inputs can approach retraining with gold labels. FaLA~\cite{fala} transfers a personalized on-device classifier to a new encoder, but selects training pairs using labels and evaluates smaller models. 
\sys examines the complementary system setting of one user's billion-parameter personal model on a phone after the labels have been discarded.

\noindent\textbf{Knowledge distillation.} 
Standard distillation matches a teacher's output distribution~\cite{hinton-kd}, while sequence-level distillation uses generated text~\cite{seq-kd}. Learning without Forgetting~\cite{lwf} uses teacher predictions to retain older behavior. Born-again networks~\cite{born-again} show that a student can match or exceed a teacher when trained on its training examples. FedMD~\cite{fedmd} and FedDF~\cite{feddf} distill across model architectures using shared public data, and data-free distillation~\cite{dafl} uses generated inputs. Universal logit distillation~\cite{uld} compares sorted next-token distributions across tokenizers. \sys applies these ideas at task-answer level: label probabilities for closed-set tasks and generated text for open-ended tasks, so the two base models need not share a vocabulary.

\noindent\textbf{On-device personalization.} 
EchoPFL~\cite{echopfl} coordinates asynchronous personalization across mobile devices, and DAPPER~\cite{dapper} estimates personalized-model accuracy without labels. For language models, XPerT~\cite{xpert} selects a suitable model from other users to accelerate on-device adaptation. 
EdgeTune~\cite{edgetune} tests whether an adapter can be reused across model versions or requires retuning with retained labels. Its weight-based test presumes comparable parameter shapes.
Phones have been used to fine-tune billion-parameter models~\cite{geyko-phone-ft} and perform NPU-based editing~\cite{mobiedit}. M4~\cite{m4} proposes an OS-managed foundation-model interface with task adapters and recognizes that adapters are version-specific. \sys addresses the point at which an already personalized adapter must be moved to a different base while its original labels are no longer stored.

\noindent\textbf{Regression-free model updates.} 
Backward-compatible representation learning~\cite{bct} and Positive-Congruent Training~\cite{pc-training} study how model updates can preserve useful earlier behavior. The latter introduces negative flip rate as a measure of newly incorrect predictions. MUSCLE~\cite{muscle} uses distillation with labeled data to reduce these flips, and self-synthesized rehearsal~\cite{ssr} generates additional examples for maintaining past behavior. We use negative flip rate to assess the user-facing stability of personalization after an upgrade.

\noindent\textbf{Per-user personalization of language models.} 
LaMP~\cite{lamp} and LongLaMP~\cite{longlamp} provide tasks with per-user histories for classification and generation. OPPU~\cite{oppu} combines a personal adapter with retrieval, and comparisons of retrieval with parameter-efficient fine-tuning show that their relative benefits depend on task and history size~\cite{salemi-rag-peft}. Our work focuses on a different stage of the personalization lifecycle: preserving a previously trained per-user adapter when the underlying base model is replaced.

\section{Conclusion}
\label{sec:conclusion}

A base-model upgrade can leave a phone's personalized adapter unusable, even though the old personal model still reflects the user's preferences. \sys preserves this information by recording the old model's predictions and training an adapter for the new base to reproduce them, without retaining labels or loading both bases into memory together. Across three generation upgrades and an architecture-preserving refresh, stored-input migration maintains the old model's mean accuracy on movie tagging and performs close to labeled retraining. On product rating, access to more of the unlabeled history improves the result, while on headline generation, distillation preserves the old model's performance but does not match the further gains from retraining with labels. On the OnePlus 15, a fixed-length 4B migration takes 38 minutes and consumes 20\% of the battery end to end. The results also make the trade-off clear: retaining fewer inputs can reduce migration accuracy or require more device computation. 
These results demonstrate a practical path for preserving on-device personalization across base-model generations, even when the user's original labels are no longer available.

\bibliographystyle{trustai4s-template/mlsys2025}
\bibliography{references}

\clearpage
\renewcommand{\labheadsuffix}{: Appendix}
\appendix

\section{Training-length selection for generation}
\label{sec:app-gen}

Table~\ref{tab:lamp4-grid} reports the training-length results for headline generation and the retraining grid. Unlike tagging, generation does not use a separate validation run for each user. Instead, we select the retraining configuration that achieves the highest mean ROUGE-L on the other nine users. This procedure selects one epoch at a learning rate of 10-4 on both upgrades. Selecting a configuration on each user's own test set would improve the score by only 0.24 and 0.14 points, compared with 3.0 and 1.6 points on tagging. This smaller gain is consistent with the lower run-to-run variation in ROUGE-L: two retraining runs at 10-4 for two epochs differ by 0.20 and 0.26 points per user on the two upgrades, compared with 2.35 and 2.99 points per user and fold on tagging.

\begin{table}[ht]
\caption{Headline generation, ROUGE-L F1 $\times$100 over 10 users at every training length (epochs). Distillation
rows show the run on all prompts at each checkpoint, with a superscript counting the users for whom held-out selection
picked that length. Retraining rows show every learning rate and length run. lr denotes learning rate.}
\label{tab:lamp4-grid}
\footnotesize
\setlength{\tabcolsep}{4pt}
\begin{tabular}{@{}lcccccc@{}}
\toprule
Row & Pair & 1 & 2 & 4 & 8 & 16 \\
\midrule
\sys, 512 stored & 2B & 12.31$^{10}$ & 12.37 & 12.18 & 12.18 & 12.32 \\
 & 4B & 13.63$^{10}$ & 13.64 & 13.51 & 13.76 & 13.35 \\
\sys, 128 stored & 2B & 11.52$^{8}$ & 12.03$^{2}$ & 11.74 & 11.79 & 11.67 \\
 & 4B & 13.34$^{9}$ & 13.44$^{1}$ & 12.75 & 12.73 & 12.98 \\
\sys, public & 2B & 12.12$^{10}$ & 11.74 & 11.56 & 11.40 & 11.17 \\
 & 4B & 13.33$^{10}$ & 12.63 & 12.88 & 12.85 & 12.53 \\
Non-personal teacher & 2B & 10.89$^{9}$ & 10.72$^{1}$ & 10.53 & 10.62 & 10.66 \\
 & 4B & 12.27$^{10}$ & 12.43 & 12.13 & 12.16 & 12.46 \\
\midrule
Retraining, lr $5{\times}10^{-5}$ & 2B & 12.89 & 12.94 &  &  &  \\
 & 4B & 14.36 & 14.01 &  &  &  \\
Retraining, lr $10^{-4}$ & 2B & 13.11 & 12.45 & 11.01 & 11.21 &  \\
 & 4B & 14.62 & 13.85 & 12.43 & 12.71 &  \\
Retraining, lr $2{\times}10^{-4}$ & 2B &  & 12.15 & 10.74 & 10.87 &  \\
 & 4B &  & 13.37 & 12.07 & 12.35 &  \\
Retraining, lr $5{\times}10^{-4}$ & 2B &  & 10.44 & 9.59 & 9.39 &  \\
 & 4B &  & 11.49 & 9.65 & 8.88 &  \\
\bottomrule
\end{tabular}

\end{table}

Short training is generally preferable for headline generation. In an initial grid of 2, 4, 8, and 16 epochs, held-out selection chose the shortest length for every user, so we added a one-epoch checkpoint. With this expanded grid, one epoch was selected for 28 and 29 of the 30 user–prompt-set combinations on the two upgrades. Each user contributes three prompt sets: up to 512 stored inputs, 128 stored inputs, and 512 public prompts. Retraining at 10-4 loses 2.1 and 2.2 ROUGE-L points when extended from one to four epochs. The recipe used to train the original personal adapter (two epochs at $2\times10^{-4}$) performs worse after the upgrade, scoring 12.15 and 13.37 compared with the pre-upgrade scores of 12.66 and 13.41.

The effect of training length depends on the prompt source. One epoch gives the highest mean score among the five checkpoints for public prompts on both upgrades. With 512 stored inputs, scores vary by at most 0.2 and 0.4 points across one to 16 epochs. With 128 stored inputs, two epochs outperform one on both upgrades. In particular, on the 2B upgrade, held-out selection chooses one epoch for most users even though its mean score is 11.52, below the 12.03 achieved at two epochs. Thus, while held-out selection usually favors short training on generation, it does not always identify the best length for a smaller stored-input set.

\section{Universal logit distillation on headline generation}
\label{sec:app-uld}

\begin{table}[t]
\caption{Universal logit distillation (ULD) on headline generation, ROUGE-L F1 $\times$100, 10 users. The rows are \sys's
text distillation from at most 512 stored inputs, rerun for this comparison, and the same with the ULD term at weight
$\lambda$. $\Delta$ is the difference to the old personal model and, with its 95\% interval, to the raw-text row. Epochs
chosen counts the users for whom held-out selection picked each length.}
\label{tab:uld}
\footnotesize
\setlength{\tabcolsep}{4pt}
\begin{tabular}{@{}lrrrw{c}{1.4em}w{c}{1.4em}w{c}{1.4em}w{c}{1.4em}w{c}{1.4em}@{}}
\toprule
 & & & & \multicolumn{5}{c}{Epochs chosen (users)} \\
\cmidrule(l){5-9}
Method & R-L & $\Delta$ & $\Delta$ to raw text [95\% CI] & 1 & 2 & 4 & 8 & 16 \\
\midrule
\multicolumn{9}{@{}l}{\emph{2B upgrade, Qwen3-1.7B $\rightarrow$ Qwen3.5-2B}} \\
Raw text, rerun & 12.25 & \textminus 0.41 & -- & 10 & 0 & 0 & 0 & 0 \\
+ ULD, $\lambda{=}1.5$, $k{=}1024$ & 12.50 & \textminus 0.16 & +0.25 [\textminus 0.39, +0.91] & 1 & 8 & 1 & 0 & 0 \\
+ ULD, $\lambda{=}0.5$, $k{=}1024$ & 12.44 & \textminus 0.23 & +0.19 [\textminus 0.18, +0.61] & 6 & 4 & 0 & 0 & 0 \\
\midrule
\multicolumn{9}{@{}l}{\emph{4B upgrade, Qwen3-4B $\rightarrow$ Qwen3.5-4B}} \\
Raw text, rerun & 13.75 & +0.34 & -- & 10 & 0 & 0 & 0 & 0 \\
+ ULD, $\lambda{=}1.5$, $k{=}256$ & 13.70 & +0.29 & \textminus 0.05 [\textminus 0.54, +0.50] & 2 & 4 & 3 & 1 & 0 \\
\bottomrule
\end{tabular}

\end{table}

Table~\ref{tab:uld} compares text-only distillation with universal logit distillation (ULD)~\cite{uld} on headline generation. We add the ULD loss to the same text-distillation procedure, using the published weight $\lambda = 1.5$ and, on the 2B upgrade, also $\lambda = 0.5$. To preserve \sys's two-phase execution, phase 1 stores a truncated version of the teacher's output distribution instead of computing the full distribution online as in ULD. For each user, we retrain the teacher and rerun the text-only baseline so that the ULD variants and the baseline share the same teacher instance. The rerun differs from the main 512-input result in Table~\ref{tab:lamp4} by -0.07 and +0.11 ROUGE-L points on the two upgrades.

At each headline position, phase 1 stores the teacher's largest 256 next-token probabilities and the remaining probability mass, without token IDs. If the mean mass outside the top 256 exceeds the prespecified 5\% threshold, we instead retain the top 1,024 probabilities. On the 2B upgrade, the top 256 leave 5.9\% of the mass, so we store 1,024 values per position. The remaining mass averages 2.0\% and reaches at most 33.4\%. On the 4B upgrade, the top 256 leave 3.8\% on average and at most 55.7\%, so 256 values are retained. For at most 512 stored inputs, the resulting records occupy 18.3 and 4.6 MB per user, respectively. We treat the omitted probability mass as a single bin in the loss, obtaining a lower bound on the distance between the full sorted distributions used by online ULD.

In phase 2, the teacher's headline, including the end-of-turn token, may have different lengths under the two tokenizers. We align positions by index up to the shorter length. On average, the student's token count is 0.998 times the teacher's on both upgrades. At each aligned position, the additional loss is the L1 distance between the student's sorted next-token probabilities and the stored teacher probabilities, including the two tail bins. We add this loss, weighted by $\lambda$, to the text-distillation cross-entropy and select the training length using the student's negative log-likelihood on held-out teacher headlines. With ULD, only one and two users select a single epoch on the two upgrades. The others train longer. The additional loss requires sorting the student's full vocabulary distribution and backpropagating through the sort at each output position, which the phone's NPU training engine does not support. So we report ULD as a GPU-only accuracy comparison, without an on-device cost measurement.

\section{Sensitivity to the vendor's adapter}
\label{sec:app-prior}
\label{sec:eval-prior}

One vendor adapter is trained with the recipe of the old base's vendor adapter, the other is tuned for accuracy on a
public validation split. Every method on the new base starts from one of them. On the first fold of the 2B and 4B
upgrades, the tuned adapter is the weaker of the two for the users.

The starting vendor adapter affects the accuracy of every subsequent method. On the first fold of the 2B upgrade, the adapter that is 7.4 points weaker for the evaluated users also lowers \sys with stored inputs by 1.6 points and labeled retraining by 2.2 points relative to the old personal model. On the 4B upgrade, the weaker starting adapter likewise reduces the accuracy of both methods. 
Tables~\ref{tab:prior} and~\ref{tab:prior-second} show that the \sys–retraining difference does not widen in these first-fold comparisons. It changes from -1.0 to -0.3 points on the 2B upgrade and from -1.4 to -0.6 points on the 4B upgrade.
This suggests that the choice of vendor adapter affects the absolute performance of both methods more than their relative difference in these tests.

\begin{table}[ht]
\begin{minipage}[t]{0.49\textwidth}
\caption{Two vendor adapters on the 2B upgrade, first fold, 19 users. $\Delta$ is the difference to the old personal
model. Retraining uses the fixed recipe.}
\label{tab:prior}
\centering
\footnotesize
\setlength{\tabcolsep}{3pt}
\begin{tabular}{@{}lrrrr@{}}
\toprule
 & \multicolumn{2}{c}{Old recipe} & \multicolumn{2}{c}{Tuned on public} \\
\cmidrule(lr){2-3}\cmidrule(l){4-5}
Method & Acc. & $\Delta$ & Acc. & $\Delta$ \\
\midrule
Vendor's adapter & 36.2 & \textminus 18.9 & 28.8 & \textminus 26.1 \\
Retraining & 55.2 & 0.0 & 52.7 & \textminus 2.2 \\
\sys, stored & 54.3 & \textminus 0.9 & 52.4 & \textminus 2.5 \\
\sys, public & 50.8 & \textminus 4.4 & 51.3 & \textminus 3.6 \\
\sys, self-generated & 52.5 & \textminus 2.7 & 49.8 & \textminus 5.1 \\
\bottomrule
\end{tabular}

\end{minipage}\hfill
\begin{minipage}[t]{0.48\textwidth}
\caption{The same comparison on the 4B upgrade, first fold, the \nfourSensUsers{} users who have every method under
both adapters. The last row is \sys with stored inputs minus retraining.}
\label{tab:prior-second}
\centering
\footnotesize
\setlength{\tabcolsep}{3pt}
\begin{tabular}{@{}lcc@{}}
\toprule
$\Delta$ & Old recipe & Tuned on public \\
\midrule
Vendor's adapter & \textminus 13.0 & \textminus 20.2 \\
Retraining & +2.5 & \textminus 2.9 \\
\sys, stored & +1.1 & \textminus 3.5 \\
\sys, public & \textminus 1.1 & \textminus 3.0 \\
\sys, self-generated & \textminus 2.7 & \textminus 5.1 \\
\midrule
Stored $-$ retraining & \textminus 1.4 & \textminus 0.6 \\
\bottomrule
\end{tabular}

\end{minipage}
\end{table}

\section{Retraining grids}
\label{sec:app-p0grid}

\begin{table}[t]
\caption{Retraining with labels at every cell of its grid, accuracy $\times$100. Val.\ is the pooled validation
accuracy of the first fold's users, each trained on four fifths of the history and scored on the rest, and Test the
accuracy over all four folds, averaged over users. The asterisk marks the fixed recipe
(ties go to fewer epochs, then the lower rate).}
\label{tab:p0grid}
\footnotesize
\setlength{\tabcolsep}{4pt}
\begin{tabular}{@{}llrrrrrrrrrr@{}}
\toprule
 & & \multicolumn{2}{c}{1 epoch} & \multicolumn{2}{c}{2 epochs} & \multicolumn{2}{c}{4 epochs} & \multicolumn{2}{c}{8 epochs} & \multicolumn{2}{c}{16 epochs} \\
\cmidrule(lr){3-4}\cmidrule(lr){5-6}\cmidrule(lr){7-8}\cmidrule(lr){9-10}\cmidrule(l){11-12}
Upgrade & Rate & Val. & Test & Val. & Test & Val. & Test & Val. & Test & Val. & Test \\
\midrule
\multicolumn{12}{@{}l}{\emph{Movie tagging, retraining with labels}} \\
Qwen3-1.7B $\rightarrow$ Qwen3.5-2B & $5{\times}10^{-5}$ &  &  & 57.4 & 58.1 & 57.7 & 57.0 & \cellcolor{black!6}57.9\rlap{$^{\ast}$} & \cellcolor{black!6}56.7 & 57.9 & 56.9 \\
 & $10^{-4}$ &  &  & 54.6 & 54.5 & 52.8 & 55.7 & 55.6 & 56.1 & 54.9 & 56.0 \\
 & $2{\times}10^{-4}$ &  &  & 48.7 & 52.0 & 49.7 & 52.1 & 54.1 & 52.6 &  &  \\
 & $5{\times}10^{-4}$ &  &  & 37.4 & 39.3 & 36.9 & 39.4 & 36.9 & 38.8 &  &  \\
Qwen3-4B $\rightarrow$ Qwen3.5-4B & $5{\times}10^{-5}$ &  &  & \cellcolor{black!6}61.0\rlap{$^{\ast}$} & \cellcolor{black!6}61.2 & 60.0 & 59.8 & 60.8 & 60.1 & 60.5 & 59.9 \\
 & $10^{-4}$ &  &  & 56.9 & 57.7 & 56.4 & 58.7 & 59.2 & 59.8 & 59.0 & 59.6 \\
 & $2{\times}10^{-4}$ &  &  & 55.4 & 53.8 & 54.4 & 53.2 & 56.2 & 54.6 &  &  \\
 & $5{\times}10^{-4}$ &  &  & 39.5 & 41.0 & 38.5 & 42.4 & 42.1 & 38.3 &  &  \\
Phi-3.5-mini $\rightarrow$ Phi-4-mini & $5{\times}10^{-5}$ &  &  & 57.9 & 57.4 & 58.2 & 57.5 & 59.0 & 57.4 & 58.7 & 57.4 \\
 & $10^{-4}$ &  &  & 57.4 & 56.9 & 57.9 & 56.8 & 59.2 & 57.2 & \cellcolor{black!6}60.5\rlap{$^{\ast}$} & \cellcolor{black!6}57.2 \\
 & $2{\times}10^{-4}$ &  &  & 55.4 & 54.4 & 52.3 & 53.3 & 55.1 & 55.2 & 55.1 & 55.7 \\
\midrule
\multicolumn{12}{@{}l}{\emph{Product rating, retraining with labels}} \\
Qwen3-4B $\rightarrow$ Qwen3.5-4B & $2{\times}10^{-5}$ & \cellcolor{black!6}76.8\rlap{$^{\ast}$} & \cellcolor{black!6}76.8 & 75.7 & 75.9 &  &  &  &  &  &  \\
 & $5{\times}10^{-5}$ & 76.3 & 76.6 & 75.8 & 75.4 & 75.1 & 76.4 & 75.9 & 76.8 &  &  \\
 & $10^{-4}$ & 73.6 & 75.0 & 74.4 & 75.1 & 74.1 & 75.4 & 74.6 & 74.6 &  &  \\
Phi-3.5-mini $\rightarrow$ Phi-4-mini & $2{\times}10^{-5}$ & 75.5 & 76.2 & 76.9 & 76.4 & 76.1 & 76.2 & 75.6 & 76.7 & 75.7 & 76.6 \\
 & $5{\times}10^{-5}$ & 76.6 & 76.8 & 75.7 & 76.0 & 76.8 & 76.0 & 77.1 & 76.5 & \cellcolor{black!6}77.7\rlap{$^{\ast}$} & \cellcolor{black!6}77.0 \\
 & $10^{-4}$ & 74.1 & 75.9 & 75.7 & 74.9 & 76.6 & 75.6 & 75.5 & 76.2 & 76.7 & 77.2 \\
\midrule
\multicolumn{12}{@{}l}{\emph{Product rating, retraining with labels on the 128 stored inputs}} \\
Qwen3-4B $\rightarrow$ Qwen3.5-4B & $2{\times}10^{-5}$ & 73.5 & 74.7 & \cellcolor{black!6}75.4\rlap{$^{\ast}$} & \cellcolor{black!6}75.2 &  &  &  &  &  &  \\
 & $5{\times}10^{-5}$ & 72.9 & 74.7 & 74.8 & 75.1 & 72.5 & 73.8 & 74.0 & 75.2 &  &  \\
 & $10^{-4}$ & 72.7 & 74.3 & 71.3 & 72.9 & 72.9 & 73.7 & 73.3 & 73.9 &  &  \\
Phi-3.5-mini $\rightarrow$ Phi-4-mini & $2{\times}10^{-5}$ & 74.0 & 74.6 & 75.0 & 75.3 & 74.6 & 75.7 & 75.0 & 75.1 & 74.8 & 75.4 \\
 & $5{\times}10^{-5}$ & 73.7 & 74.2 & 76.0 & 75.5 & 75.8 & 74.7 & 76.9 & 75.0 & \cellcolor{black!6}78.1\rlap{$^{\ast}$} & \cellcolor{black!6}75.4 \\
 & $10^{-4}$ & 71.2 & 73.2 & 75.0 & 74.0 & 73.7 & 73.6 & 75.2 & 74.1 & 75.8 & 75.1 \\
\bottomrule
\end{tabular}

\end{table}

Table~\ref{tab:p0grid} gives every cell of the retraining grids, for the \ntwoUsers{} users of movie tagging and the
first \nlthreeUsers{} users of product rating. On product rating, retraining on the 128 stored inputs has its own
recipe, chosen the same way. It is \nlthreeRetrainItemsOneTwoEightRecipe{} on the 4B upgrade and
\nlthreePhiRetrainItemsOneTwoEightRecipe{} on the Phi upgrade.

Retraining the new base with the personal adapter's original recipe (two epochs at $2\times10^{-4}$) does not recover the pre-upgrade tagging accuracy. Its differences from the old personal model are -2.3 [-4.8, +0.1], -5.1 [-6.8, -3.6], and -2.7 [-4.8, -0.7] points on the 2B, 4B, and Phi upgrades. Extending training to four or eight epochs does not consistently recover this loss. For example, on the 4B upgrade, the differences remain -5.6 [-7.6, -3.9] and -4.3 [-6.9, -1.7] points at four and eight epochs. On Phi, 16 epochs reduces the loss to -1.3 [-3.2, +0.7] points. In contrast, the validation-selected fixed recipes in Table~\ref{tab:p0grid} use learning rates below $2\times10^{-4}$ on all three upgrades, and each outperforms the corresponding $2\times10^{-4}$ configuration. 

We expanded each retraining grid when its validation-selected configuration lay on the boundary of the initial search range. Table~\ref{tab:p0grid} reports all evaluated configurations and marks the selected recipes. The extensions led to lower selected learning rates on the two Qwen tagging upgrades and the 4B product-rating upgrade.

For Phi product rating, the selected 16-epoch recipe has a 0.7-point validation advantage over eight epochs, comparable to run-to-run variation. The two settings tie in the extension run, and choosing eight epochs would lower the reported retraining accuracy by 0.5 points without changing the interpretation. Because the selected configuration lies outside the initial grid, the four-fold Phi retraining results for both full-history and 128-input settings come from the extension run, using the same implementation and vendor-adapter initialization. Where the runs overlap, per-user-fold accuracies differ by 1.62 and 1.05 points on average for the two settings, respectively.

Per-user tuning selects from 14, 14, and 12 grid settings on the three tagging upgrades, and from 10 and 15 settings on product rating. Because each choice is based on one user's limited data, it is sensitive to both individual differences and training noise. On tagging, its difference from the fixed recipe is +0.3 [-1.5, +2.1], -2.2 [-3.6, -0.9], and +0.5 [-1.3, +2.4] points across the three upgrades. For comparison only, choosing the best setting retrospectively on each user's test items would add 2.7, 3.8, and 2.8 points over per-user tuning.

\section{Adapter rank}
\label{sec:app-rank}

\begin{table}[t]
\caption{Adapter rank on the 2B upgrade, first fold, \nrankSixteenUsers{} users, accuracy $\times$100. Retraining uses
the rank-16 and the re-selected recipe, and \sys trains at $2\times10^{-4}$ and at the rate held-out selection picks.
$\Delta$ is \sys at the selected rate minus retraining, with its 95\% interval.}
\label{tab:rank}
\footnotesize
\setlength{\tabcolsep}{4pt}
\begin{tabular}{@{}lccc@{}}
\toprule
 & \multicolumn{3}{c}{Adapter rank} \\
\cmidrule(l){2-4}
Method & 8 & 16 & 32 \\
\midrule
Vendor's adapter & 36.5 & 36.2 & 34.5 \\
Before the upgrade & 54.9 & 55.2 & 54.9 \\
Retraining, rank-16 recipe & 53.9 & 55.2 & 56.4 \\
Retraining, re-selected recipe & 54.6 & 55.2 & 56.0 \\
\sys, stored inputs, at $2{\times}10^{-4}$ & 56.6 & 54.3 & 50.3 \\
\sys, stored inputs, at the selected rate & 54.9 & 54.5 & 55.9 \\
\midrule
\multicolumn{4}{@{}l}{\emph{\sys at the selected rate $-$ retraining}} \\
Rank-16 recipe, $\Delta$ [95\% CI] & +1.0 [\textminus 2.2, +3.9] & \textminus 0.7 [\textminus 4.3, +2.9] & \textminus 0.6 [\textminus 3.2, +2.0] \\
\quad up/down & 10/5 & 8/7 & 7/8 \\
Re-selected recipe, $\Delta$ [95\% CI] & +0.3 [\textminus 2.6, +3.4] & \textminus 0.7 [\textminus 4.3, +2.9] & \textminus 0.2 [\textminus 2.1, +2.0] \\
\quad up/down & 7/9 & 8/7 & 5/7 \\
\bottomrule
\end{tabular}

\end{table}

Table~\ref{tab:rank} reports adapter-rank sensitivity on the 2B upgrade.
At each rank $r$, the vendor adapter, retraining baseline, and \sys student all use rank $r$, with LoRA alpha set to 2$r$. Retraining is evaluated using the rank-16 recipe (eight epochs at $5{\times}10^{-5}$) and a recipe reselected for each rank from $\{5{\times}10^{-5}, 10^{-4}, 2{\times}10^{-4}\} \times \{2, 4, 8, 16\}$ epochs on first-fold validation data. Ties favor fewer epochs and then the smaller learning rate. The reselected recipes are eight epochs at $10^{-4}$ for rank 8 and four epochs at $5{\times}10^{-5}$ for rank 32.

\sys distills from at most 128 stored inputs using epoch candidates $\{2, 4, 8, 16\}$ and learning rates $\{5{\times}10^{-5}, 10^{-4}, 2{\times}10^{-4}, 5{\times}10^{-4}, 10^{-3}\}$, with one teacher per user. The epoch count is selected per user. We select one learning rate per rank using the mean held-out KL divergence, across users, between the teacher's and student's distributions over complete label strings. Unlike the first-token KL used to select epochs, this criterion captures changes in later tokens of multi-token labels, which also affect the full-sequence likelihood used at inference. Rates within 0.02 of the lowest mean KL are treated as tied. We prefer $2{\times}10^{-4}$ when it is among them, otherwise the smallest tied rate.

The $2{\times}10^{-4}$ row comes from the earlier experiment at each rank (from the main first fold at rank 16). 
At ranks 8 and 32, selected-rate \sys and reselected retraining are separate runs initialized from the same vendor adapter. At rank 16, the retraining rows use the main run.
Table~\ref{tab:rank} reports \sys minus retraining with user-bootstrap 95\% intervals, together with the number of users for whom each method scores higher or lower.

Across ranks 8, 16, and 32, \sys with a selected learning rate remains within 0.7 points of the re-tuned retraining baseline on the first fold. In contrast, retaining the original distillation learning rate substantially reduces accuracy at rank 32. These results suggest that the learning rate matters more than adapter rank within the tested range.

\section{LoRASuite on the Phi upgrade}
\label{sec:app-lorasuite}

\begin{table}[t]
\caption{LoRASuite mapping on the Phi upgrade (19 users, four folds). Accuracy and negative flip rates are reported before and after fine-tuning with up to 100 labeled inputs. Fine-tuning starts from the mapped adapter, the vendor adapter, or a fresh LoRA. $\Delta$ and its 95\% confidence interval are relative to the old personal model.}
\label{tab:lorasuite}
\footnotesize
\setlength{\tabcolsep}{4pt}
\begin{tabular}{@{}lccc@{}}
\toprule
Method & Acc. & $\Delta$ [95\% CI] & Flips \\
\midrule
No adapter & 26.3 & \textminus 30.8 [\textminus 36.9, \textminus 25.6] & 62 \\
Vendor's adapter & 43.3 & \textminus 13.7 [\textminus 19.2, \textminus 8.8] & 38 \\
Mapped adapter & 8.9 & \textminus 48.1 [\textminus 55.3, \textminus 41.0] & 90 \\
\midrule
Mapped, then fine-tuned on 100 labels & 26.2 & \textminus 30.8 [\textminus 34.8, \textminus 27.1] & 66 \\
Vendor's adapter, the same fine-tuning & 57.7 & +0.7 [\textminus 1.1, +2.5] & 16 \\
Fresh LoRA, the same fine-tuning & 47.6 & \textminus 9.4 [\textminus 11.9, \textminus 7.1] & 33 \\
\bottomrule
\end{tabular}

\end{table}

Table~\ref{tab:lorasuite} 
provides the detailed LoRASuite results discussed in Section~\ref{sec:eval-samearch}.
The paper's transformation
multiplies each head's update by the two bases' projection weights. 
On this model pair, the transformation substantially enlarges the adapter updates, especially in the first layer, and the mapped adapter performs worse than the new base without an adapter.

For \nphiLsUsers{} users and \nphiLsFolds{} folds, each fold's old personal model, the main run's teacher, is trained
again inside this run with the main run's recipe. Its adapter is then mapped onto Phi-4-mini by our implementation of
LoRASuite's mapping~\cite{lorasuite}, written from the paper's equations. The three fine-tuned rows follow the paper's
lightweight fine-tuning on the last 100 labeled items of the fold's training history (the whole history for the
\nphiLsShortFolds{} of \nphiLsUserFolds{} user-folds that have fewer). Each runs 3 epochs at batch 16,
\nphiLsStepsMin{} to \nphiLsStepsMax{} steps, with the learning rate decaying linearly to zero and no warm-up. Of the
paper's two published rates, \nphiLsRate{} gives the mapped start the higher mean and is used for all three rows.
At \nphiLsRateOther{} they end at \nphiLsLftAltAcc{}, \nphiLoraHundredAltAcc{} and \nphiLoraHundredFreshAltAcc{}, in
the table's order. The three rows share one teacher instance, so their comparisons with the main run's rows are
across runs, like those of the rows started from the partial copy of the old adapter.

We implement the mapping by aligning layers, attention heads, and hidden dimensions, then transferring each adapter's dense update $\Delta W = sBA$. Because the two Phi models have the same number of layers, the layer alignment is the identity. Their hidden widths are equal, but their vocabularies differ. 
We retain the hidden-space map $W_h$, estimated by least squares from the embeddings of the 28,701 tokens whose byte strings occur in both vocabularies.
On this token intersection, the old embedding matrix has condition number 15 and the map has a relative residual of 48.8\%.

Following the paper, we match attention heads using cosine similarity between their query–key and value–output products, transforming the old products through $W_h$. We solve the padded assignment using the Hungarian algorithm. Of the 32 old heads per layer, eight match padding entries and are dropped, accounting for 24.8\% of the squared query–key–value update on average. Each of the eight new key–value heads receives the mean transformed key and value updates from its three matched query heads. Because the paper does not specify a separate update rule for feed-forward projections, we apply the analogous attention mapping with an intermediate-space transform $W_i$ derived from the projection weights. Each mapped update remains within the original adapter rank, and the singular value decomposition (SVD) used to recover LoRA factors has relative error at most \nphiLsSvdErrMax{}.

The paper's per-head transformation, its Equation~3, is
\begin{equation}
(\Delta W^{j}_{Q})_{n} = W_{h}^{\top}\,(\Delta W^{i}_{Q})_{o}\,(W^{i}_{Q})_{o}^{\top}\,W_{h}\,(W^{j}_{Q})_{n}
\label{eq:lorasuite}
\end{equation}
for old head $i$ matched to new head $j$, and alike for the key, value and output projections. The weights enter
transposed, not inverted, so the transformation is not the identity even when a base is mapped onto itself. With
$W_h = I$ and $i = j$ it reduces to $\Delta W\,W^{\top}W$, the update times the head's Gram matrix. The feed-forward
update is projected onto the row space of the base's projection, 3,072 of its 8,192 dimensions. Mapped from
Phi-3.5-mini onto Phi-3.5-mini itself, one user's adapter loses \nphiLsSelfMapLoss{} points (user 1, first fold).
There the computed $W_h$ is the identity and every head is matched to itself. The query, key and value updates still
grow by a factor of \nphiLsSelfNormQkvMin{} to \nphiLsSelfNormQkvMax{} in norm across the layers. The whole mapped
adapter changes the user's accuracy by \nphiLsSelfMapVsOrig{} points (the loss above), its mapped attention updates
alone by \nphiLsSelfAttnOnlyVsOrig{} and its mapped feed-forward updates alone by \nphiLsSelfMlpOnlyVsOrig{}.
These diagnostics suggest that the published transformation is the main source of the loss in our implementation. We ran the same code with pseudo-inverses in place of the
transposes and $W_i$ set to the identity, which is not the published method. That variant maps the adapter onto
itself with a relative error of at most \nphiLsInvSelfRelErrMax{} on every module and changes \nphiLsInvSelfChanged{}
of its predictions.

On the upgrade, the first layer's query, key and value update grows by a factor of \nphiLsNormFirst{} on average over
the \nphiLsUserFolds{} adapters, and the other layers' by \nphiLsNormRestMin{} to \nphiLsNormRestMax{}. This is why
the mapped adapter ends below the new base without an adapter, and why \nphiLsStepsMax{} steps of fine-tuning at
\nphiLsRate{} do not recover it. On the first fold, the whole mapped adapter puts the \nphiLsUsers{} users at
\nphiLsMapFoldZeroAcc{}, against \nphiZeroshotFoldZeroAcc{} for the new base without an adapter on the same items.
The mapped attention updates alone put them at \nphiLsAttnOnlyAcc{}, and the mapped feed-forward updates alone at
\nphiLsMlpOnlyAcc{}.
In the paper's own Llama and Qwen pairs, the mapped adapter before fine-tuning ends within a point of the new base
without an adapter (its appendix tables). On the Phi upgrade it ends \nphiLsMapVsZeroshot{} points from it.

\section{Adapted synthetic-prompt filters}
\label{sec:app-filters}
\begin{table}[t]
\caption{Adapted synthetic-prompt filters on movie tagging (19 users, four folds). $\Delta$ old compares with the pre-upgrade personal model, while $\Delta$ unfiltered compares with unfiltered synthetic prompts. Both differences include 95\% confidence intervals.}
\label{tab:filters}
\footnotesize
\setlength{\tabcolsep}{4pt}
\begin{tabular}{@{}lcccc@{}}
\toprule
Method & Acc. & $\Delta$ old & $\Delta$ unfiltered & Flips \\
\midrule
\multicolumn{5}{@{}l}{\emph{2B upgrade, Qwen3-1.7B$\rightarrow$3.5-2B}} \\
Self-generated, unfiltered (rerun) & 52.9 & \textminus 1.4 [\textminus 3.1, +0.2] &  & 16 \\
Trans-LoRA filter & 53.1 & \textminus 1.3 [\textminus 2.8, +0.3] & +0.2 [\textminus 0.8, +1.1] & 15 \\
TuneShift-KD filter & 49.8 & \textminus 4.6 [\textminus 6.8, \textminus 2.3] & \textminus 3.1 [\textminus 4.6, \textminus 1.7] & 21 \\
\midrule
\multicolumn{5}{@{}l}{\emph{4B upgrade, Qwen3-4B$\rightarrow$3.5-4B}} \\
Self-generated, unfiltered (rerun) & 56.8 & \textminus 2.1 [\textminus 3.3, \textminus 0.9] &  & 14 \\
Trans-LoRA filter & 56.8 & \textminus 2.1 [\textminus 3.5, \textminus 0.7] & 0.0 [\textminus 1.0, +1.2] & 16 \\
TuneShift-KD filter & 53.9 & \textminus 4.9 [\textminus 6.5, \textminus 3.3] & \textminus 2.8 [\textminus 4.2, \textminus 1.4] & 21 \\
\midrule
\multicolumn{5}{@{}l}{\emph{Phi upgrade, Phi-3.5$\rightarrow$Phi-4-mini}} \\
Self-generated, unfiltered (rerun) & 54.9 & \textminus 2.1 [\textminus 4.0, \textminus 0.2] &  & 15 \\
Trans-LoRA filter & 55.5 & \textminus 1.5 [\textminus 3.1, \textminus 0.1] & +0.6 [\textminus 0.7, +2.0] & 13 \\
TuneShift-KD filter & 53.7 & \textminus 3.3 [\textminus 5.4, \textminus 1.1] & \textminus 1.2 [\textminus 2.5, +0.2] & 16 \\
\bottomrule
\end{tabular}

\end{table}

For each upgrade the new base answers \ntlPoolRequests{} requests with the recipe of its 512 self-generated prompts,
four times as many under another sampling seed. The pool holds \ntlPoolN{} distinct candidates on every upgrade, none
of them among the 512. Each filter keeps 512 of them per user and fold, and the kept prompts go through phases 1 and 2
unchanged. The unfiltered row trains on the 512 self-generated prompts of Table~\ref{tab:variants}, the first draw,
alongside the filtered rows and with the same teacher. 
Its four-fold mean
differs from the main run's by \ntwoSelfRerunVsMain{} \ntwoSelfRerunVsMainCI{}, \nfourSelfRerunVsMain{}
\nfourSelfRerunVsMainCI{} and \nphiSelfRerunVsMain{} \nphiSelfRerunVsMainCI{} points. The two Qwen differences are
training noise,
which we attribute to the GPU kernels of the Qwen3.5 bases' linear-attention
layers. Phi-4-mini, with standard attention, trains bit-reproducibly in our setup, and the rerun repeats the main
run's predictions exactly.

Trans-LoRA trains a discriminator next to the source adapter and keeps the synthetic inputs it rates as real. Our
discriminator is a fresh rank-16 adapter on the old base, trained for 2 epochs at $2{\times}10^{-4}$ on the fold's
training inputs against as many candidates drawn at random, and it ranks the remaining candidates by its probability
of real. Its training AUC averages \ntwoDiscAucTrain{}, \nfourDiscAucTrain{} and \nphiDiscAucTrain{} over users and
folds, and the AUC of the fold's test inputs against the pool \ntwoDiscAucTest{}, \nfourDiscAucTest{} and
\nphiDiscAucTest{}. Trans-LoRA writes its candidates from five seed examples of the task's data and keeps as many as
the task had, which we replace by the label names and the row's budget of 512, and its discriminator has to be trained
before the upgrade, while the user's inputs are still on the phone.

TuneShift-KD compares a fine-tuned model with the base it was fine-tuned from. It keeps a synthetic input when the
fine-tuned model gives its own answer a perplexity below 1.5 and the base's answer one of 1.5 or more. An external
model then writes further candidates from the kept ones until the set is as large as the task's data. Here the
fine-tuned model is the old personal model and its base the non-personal teacher, the old base with the vendor's
adapter, from which the personal adapter was trained.
On a closed label set an answer is a
label whose perplexity is the inverse of its probability. 
Our rule scores each candidate by the difference between the old personal model’s and the non-personal teacher’s log probabilities for the old model’s preferred label.
It keeps the 512 highest scores per user and fold in place of the two thresholds, which \ntwoGainPassShare{},
\nfourGainPassShare{} and \nphiGainPassShare{} percent of the kept prompts pass. When the two models disagree on fewer
than 512 candidates, which happens for \ntwoGainShort{}, \nfourGainShort{} and \nphiGainShort{} of \ntlUserFolds{}
user-folds, it fills the rest with candidates on which they agree, in random order. 

Both filters run in phase 1 on the old base, the discriminator as an adapter trained before the upgrade, like the personal adapter, and TuneShift-KD's rule as a second pass with the non-personal teacher. Like \sys with
self-generated prompts, they keep nothing of the history after the upgrade, and their phone cost is not measured.
These baselines adapt the published prompt-selection strategies without reproducing the original methods' complete training pipelines. The filtered configurations use a separate set of 2,048 generated candidates, while the unfiltered baseline uses the original 512 prompts. We did not evaluate random selection from the same candidate pool, so these comparisons cannot isolate the effect of filtering alone.

\section{Personalization gains before upgrades}
\label{sec:app-gate}

\begin{table}[t]
\caption{Personal gain on each old base before any upgrade, one split per user, \nbaseTaggingSmallPeople{} users
per task. The columns give accuracy $\times$100 without an adapter, with the vendor's adapter, and with per-user
adapters trained from scratch or from the vendor's adapter. $\Delta$ is the per-user adapter's difference to the
vendor's adapter, and up/down/tie counts the users above, below and level with it.}
\label{tab:gate}
\footnotesize
\setlength{\tabcolsep}{4pt}
\begin{tabular}{@{}lccccrccr@{}}
\toprule
 & & & \multicolumn{3}{c}{Per-user adapter, from scratch} & \multicolumn{3}{c}{Per-user adapter, from the vendor's} \\
\cmidrule(lr){4-6}\cmidrule(l){7-9}
Old base & No adapter & Vendor's adapter & Acc. & $\Delta$ & up/down/tie & Acc. & $\Delta$ & up/down/tie \\
\midrule
\multicolumn{9}{@{}l}{\emph{Movie tagging}} \\
Qwen3-1.7B & 21.4 & 39.2 & 45.4 & +6.2 & 13/6/1 & 52.6 & +13.3 & 15/4/1 \\
Qwen3-4B & 28.4 & 42.9 & 48.2 & +5.3 & 11/8/1 & 57.6 & +14.7 & 16/2/2 \\
Phi-3.5-mini & 31.1 & 40.8 & 48.4 & +7.6 & 12/6/2 & 55.3 & +14.6 & 16/3/1 \\
\midrule
\multicolumn{9}{@{}l}{\emph{Product rating}} \\
Qwen3-1.7B & 23.0 & 67.9 & 72.0 & +4.1 & 10/8/2 & 74.8 & +6.9 & 17/3/0 \\
Qwen3-4B & 39.2 & 63.7 & 73.3 & +9.6 & 14/6/0 & 75.3 & +11.6 & 17/2/1 \\
Phi-3.5-mini & 57.1 & 68.1 & 75.0 & +6.9 & 16/3/1 & 76.4 & +8.3 & 18/1/1 \\
\bottomrule
\end{tabular}

\end{table}

On each old base, the rule of Section~\ref{sec:eval-setup} reads the per-user adapter trained from scratch. It has to
beat the vendor's adapter for that base by at least 3 points on average and for more than half of the users, and a
tie counts as no gain. 
As shown in Table~\ref{tab:gate}, every old base passes the personalization-usefulness rule except the 1.7B base on product rating, so the 2B upgrade is evaluated on movie tagging.
Each user's split puts three quarters of the history in training and
one quarter in the test set. The per-user adapter trains from scratch for 3 epochs at $2\times10^{-4}$, or from the
vendor's adapter with the personal adapter's recipe (2 epochs). On movie tagging the users include the one excluded
from the migration results, and on product rating they are the first \nbaseRatingSmallPeople{}. 
Trained from the vendor's adapter, as the old personal models of Section~\ref{sec:eval-setup} are, per-user adapters gain more than from scratch on every base. On the 1.7B base on product rating they score \nbaseRatingSmallFromVendorVsVendor{} points against the vendor's adapter and beat it for \nbaseRatingSmallFromVendorUp{} of \nbaseRatingSmallPeople{} users.

\section{The excluded user}
\label{sec:app-user0}

\begin{table}[t]
\caption{Movie-tagging results for the excluded user with 3,190 films (three upgrades, four folds). $\Delta$ is the accuracy difference from the old personal model.}
\label{tab:user0}
\footnotesize
\setlength{\tabcolsep}{4pt}
\begin{tabular}{@{}lrrrrrr@{}}
\toprule
 & \multicolumn{2}{c}{Qwen3-1.7B $\rightarrow$ Qwen3.5-2B} & \multicolumn{2}{c}{Qwen3-4B $\rightarrow$ Qwen3.5-4B} & \multicolumn{2}{c}{Phi-3.5-mini $\rightarrow$ Phi-4-mini} \\
\cmidrule(lr){2-3}\cmidrule(lr){4-5}\cmidrule(l){6-7}
Method & Acc. & $\Delta$ & Acc. & $\Delta$ & Acc. & $\Delta$ \\
\midrule
Before the upgrade & 48.1 & -- & 51.2 & -- & 51.4 & -- \\
Retraining, fixed recipe & 46.8 & \textminus 1.3 & 51.7 & +0.6 & 48.3 & \textminus 3.1 \\
\sys, stored inputs & 48.9 & +0.8 & 48.8 & \textminus 2.4 & 51.0 & \textminus 0.4 \\
\sys, self-generated & 47.3 & \textminus 0.8 & 50.2 & \textminus 0.9 & 49.1 & \textminus 2.3 \\
Non-personal teacher & 38.1 & \textminus 10.0 & 42.5 & \textminus 8.7 & 37.6 & \textminus 13.8 \\
Vendor's adapter & 32.7 & \textminus 15.4 & 35.5 & \textminus 15.6 & 41.9 & \textminus 9.5 \\
\bottomrule
\end{tabular}

\end{table}

Table~\ref{tab:user0} gives part of Tables~\ref{tab:main} and \ref{tab:variants} for this user, run with the protocol of the other
\ntwoUsers{} users. Including this user changes the mean difference between \sys with stored inputs and the old
personal model by \ntwoUserZeroShift{}, \nfourUserZeroShift{} and \nphiUserZeroShift{} points on the three upgrades.

\section{Product rating: full results and tokenization}
\label{sec:app-rating}

\begin{table}[t]
\caption{Full product-rating results for the first 20 users on the 4B and Phi upgrades (four folds). $\Delta$ and its 95\% confidence interval are relative to the old personal model. Fixed-16-epoch rows report accuracy only. The 128-input retraining baseline uses a separately selected recipe.}
\label{tab:lamp3}
\footnotesize
\setlength{\tabcolsep}{4pt}
\begin{tabular}{@{}llccccc@{}}
\toprule
Personal & & & & & & \\
data kept & Method & Acc. & $\Delta$ & 95\% CI & Flips & MAE \\
\midrule
\multicolumn{7}{@{}l}{\emph{Qwen3-4B $\rightarrow$ Qwen3.5-4B}} \\
-- & Personal model before the upgrade & 74.9 & -- &  & -- & 0.26 \\
\addlinespace
inputs + labels & Retrieval, 8 labeled examples in the prompt & 69.4 & \textminus 5.4 & [\textminus 8.9, \textminus 2.6] & 21 & 0.32 \\
 & Retraining with labels, fixed recipe & 76.8 & +1.9 & [+0.6, +3.3] & 10 & 0.24 \\
 & Retraining with labels, per-user tuning & 76.5 & +1.6 & [+0.4, +2.9] & 11 & 0.24 \\
 & Retraining with labels, 128 stored inputs & 75.2 & +0.3 & [\textminus 0.8, +1.5] & 13 & 0.26 \\
\addlinespace
inputs & \sys, 128 stored inputs & 74.8 & \textminus 0.1 & [\textminus 1.0, +1.0] & 10 & 0.26 \\
 & \sys, all inputs & 75.8 & +0.9 & [+0.1, +1.7] & 7 & 0.25 \\
 & \sys, 128 stored inputs, fixed 16 epochs & 75.3 & +0.5 & [\textminus 0.5, +1.5] & -- & -- \\
 & \sys, all inputs, fixed 16 epochs & 76.4 & +1.6 & [+0.7, +2.5] & -- & -- \\
\addlinespace
nothing & \sys, public prompts & 74.0 & \textminus 0.9 & [\textminus 2.1, +0.2] & 10 & 0.27 \\
 & \sys, self-generated prompts & 70.0 & \textminus 4.8 & [\textminus 6.9, \textminus 3.1] & 16 & 0.31 \\
\addlinespace
-- & Non-personal teacher & 62.6 & \textminus 12.2 & [\textminus 19.3, \textminus 6.4] & 30 & 0.40 \\
 & New base with the vendor's adapter & 63.4 & \textminus 11.5 & [\textminus 16.8, \textminus 6.9] & 29 & 0.39 \\
 & New base, no adapter & 62.9 & \textminus 12.0 & [\textminus 17.1, \textminus 7.3] & 31 & 0.40 \\
\midrule
\multicolumn{7}{@{}l}{\emph{Phi-3.5-mini $\rightarrow$ Phi-4-mini}} \\
-- & Personal model before the upgrade & 76.4 & -- &  & -- & 0.24 \\
\addlinespace
inputs + labels & Retrieval, 8 labeled examples in the prompt & 68.8 & \textminus 7.7 & [\textminus 11.6, \textminus 4.4] & 22 & 0.33 \\
 & Retraining with labels, fixed recipe & 77.0 & +0.6 & [\textminus 0.4, +1.5] & 10 & 0.24 \\
 & Retraining with labels, per-user tuning & 77.1 & +0.7 & [\textminus 0.5, +1.7] & 10 & 0.24 \\
 & Retraining with labels, 128 stored inputs & 75.4 & \textminus 1.0 & [\textminus 2.3, +0.1] & 12 & 0.25 \\
\addlinespace
inputs & \sys, 128 stored inputs & 75.5 & \textminus 1.0 & [\textminus 2.2, +0.1] & 10 & 0.25 \\
 & \sys, all inputs & 76.5 & +0.1 & [\textminus 0.9, +1.0] & 7 & 0.24 \\
 & \sys, 128 stored inputs, fixed 16 epochs & 76.0 & \textminus 0.4 & [\textminus 1.5, +0.6] & -- & -- \\
 & \sys, all inputs, fixed 16 epochs & 77.5 & +1.0 & [+0.4, +1.6] & -- & -- \\
\addlinespace
nothing & \sys, public prompts & 75.0 & \textminus 1.4 & [\textminus 2.4, \textminus 0.5] & 10 & 0.26 \\
 & \sys, self-generated prompts & 72.4 & \textminus 4.0 & [\textminus 6.1, \textminus 2.4] & 14 & 0.28 \\
\addlinespace
-- & Non-personal teacher & 68.2 & \textminus 8.2 & [\textminus 12.4, \textminus 4.7] & 23 & 0.33 \\
 & New base with the vendor's adapter & 64.1 & \textminus 12.3 & [\textminus 19.3, \textminus 6.7] & 28 & 0.38 \\
 & New base, no adapter & 56.2 & \textminus 20.3 & [\textminus 26.3, \textminus 14.9] & 39 & 0.48 \\
\bottomrule
\end{tabular}

\end{table}

\begin{table}[t]
\caption{\sys's accuracy $\times$100 on product rating after each number of training epochs, for the first
\nlthreeUsers{} users over four folds. A superscript counts the user-folds whose held-out selection picked that length.
The 4B upgrade is Qwen3-4B $\rightarrow$ Qwen3.5-4B and the Phi upgrade Phi-3.5-mini $\rightarrow$ Phi-4-mini.}
\label{tab:lamp3-grid}
\footnotesize
\setlength{\tabcolsep}{4pt}
\begin{tabular}{@{}lccccc@{}}
\toprule
Row & Pair & 2 & 4 & 8 & 16 \\
\midrule
\sys, 128 stored & 4B & 73.0$^{13}$ & 73.9$^{21}$ & 75.1$^{20}$ & 75.3$^{26}$ \\
 & Phi & 74.2$^{13}$ & 75.3$^{26}$ & 75.8$^{17}$ & 76.0$^{24}$ \\
\sys, all stored & 4B & 74.4$^{15}$ & 75.7$^{18}$ & 76.3$^{18}$ & 76.4$^{29}$ \\
 & Phi & 75.6$^{22}$ & 77.0$^{12}$ & 77.3$^{22}$ & 77.5$^{24}$ \\
\sys, public & 4B & 73.2$^{3}$ & 74.0$^{17}$ & 74.1$^{26}$ & 74.0$^{34}$ \\
 & Phi & 74.5$^{6}$ & 74.7$^{13}$ & 75.2$^{33}$ & 74.9$^{28}$ \\
\sys, self-generated & 4B & 68.3$^{1}$ & 69.2$^{12}$ & 69.6$^{27}$ & 70.1$^{40}$ \\
 & Phi & 71.6 & 72.1$^{17}$ & 72.4$^{29}$ & 72.3$^{34}$ \\
Non-personal teacher & 4B & 62.7$^{3}$ & 61.9$^{13}$ & 62.3$^{22}$ & 62.7$^{42}$ \\
 & Phi & 65.6$^{8}$ & 68.7$^{14}$ & 67.7$^{24}$ & 67.9$^{34}$ \\
\bottomrule
\end{tabular}

\end{table}

Table~\ref{tab:lamp3} gives every method for the first \nlthreeUsers{} users on the 4B and Phi upgrades,
and Table~\ref{tab:lamp3-grid} gives every training length of the distillation rows on product rating.
Merged by person with the same rule as movie tagging, the user split's training and development entries give
\nlampthreePeople{} users, \nlampthreePeopleAtLeastMin{} with at least \nlampthreeMin{} reviews. We rank users
by review count and evaluate the top \nlampthreeTop{}, skipping the \nlampthreeExcluded{} who give one score to more
than \nlampthreeExcludeMajority\% of their reviews (each at least \nlampthreeExcludedMinShare\% five-star). Each user
keeps \nlampthreeCap{} random reviews, fewer than any evaluated user has (\nlampthreeTopMin{} to \nlampthreeTopMax{}),
so every history is truncated, and each review is cut to \nlampthreeMaxChars{} characters. The new base generates its
self-generated prompts, here reviews, at temperature 1, and duplicates are removed. A review carries one score, so
there are no label pairs to cover as there are for tags. Accuracy is exact match, and mean absolute error is
computed per user with its interval.

The prompt is LaMP's instruction and the answer one digit, with a distinct first token per score in every Qwen
tokenizer and Phi-4-mini's. Phi-3.5's tokenizer places a lone space token before a digit, also after the assistant marker, so every rating begins
with the same token, which the model emits with probability one. We treat it as the last prompt token and train and
score on the digit that follows (Phi-4-mini encodes each digit as one token). On movie tagging the same tokenizer
joins the space marker to the first piece of each tag, so the 15 tags already differ in their first token after the
assistant marker, and the token sequences every row trains and scores on match the joint tokenization of prompt and
tag. Both cases follow one rule, under which the answer position is the first position at which the labels' tokens
differ. For every other model and label set here, that position is the first token of the answer.

\section{Device measurements and configurations}
\label{sec:app-energy}

\textbf{Measurement.} The minutes in Table~\ref{tab:cost} are measured on the charger, screen off, with an external
cooling fan clipped to the phone. Energy is measured on battery at minimum screen brightness, fan on, by the battery's
fuel gauge (the kernel's charge counter, \nCostFuelGaugeUpdates{} updates over the session), idle draw subtracted. The
load is \nCostLoadWMin{} to \nCostLoadWMax{}\,W above idle. Throughout the paper a share of the battery is charge
drawn, by the counter, over the gauge's full charge, \nCostBatteryMAh{} mAh on the OnePlus 15 and
\nCostThirteenBatteryMAh{} on the 13T. Each is the median of counter over displayed level in that phone's energy
session. Energy in that table is the run's tokens times per-token rates, in kJ and as a share of the full charge
(\nCostBatteryMAh{} mAh). Its two rows marked measured end to end are the gauge's reading of one run, idle draw and
fixed overheads included. Peak memory is the trainer process's peak resident memory.

By the per-token rates, the 4B upgrade's migration from stored inputs takes
\nCostRateModelFixedPct\% of the battery at a fixed length, \nCostRateModelSelPct\% with selection, and retraining
\nCostRateModelRetrainPct\%. Measured end to end on battery, from the start of the teacher pass until the next job
begins, with the idle draw and the fixed overheads included, the migration takes \nCostEndMin{} minutes and draws
\nCostEndmAh{} mAh, \nCostEndkJ{}\,kJ and \nCostEndPct\% of the battery. The labeled retraining that followed drew
\nCostRetrainEndmAh{} mAh, \nCostRetrainEndkJ{}\,kJ and \nCostRetrainEndPct\%. The run starts at a displayed level of
\nCostEndPctFrom\%, where charge protection stops charging. The displayed level is an integer that lags the counter. It
falls by \nCostEndLevelDrop{} points over the migration, against the counter's \nCostEndPct{}. The measured energy is
\nCostOverheadPct\% above what the per-token rates and the idle draw predict for the same run, the fixed overheads of
starting the engine's NPU process, loading the base and cooling between jobs. The selection run scores the held-out
fifth at its four checkpoints. It takes \nCostSelExtraPctFour\% longer than its tokens would at the fixed 8-epoch run's
pace on the 4B upgrade and \nCostSelExtraPctTwo\% on the 2B. On battery the jobs run \nCostBatteryVsChargerPctMin{} to
\nCostBatteryVsChargerPctMax\% faster than on the charger, and the end-to-end run's jobs at \nCostBridgeMin{} to
\nCostBridgeMax{} times the speed of the session that measured the rates. Without the fan the fixed 8-epoch training
takes \nCostNoFanSec{} instead of \nCostNoFanFanSec{}\,s, \nCostNoFanSlowerPct\% longer, and completes. The NPU peaks
at \nCostNoFanNPUMax{} instead of \nCostNoFanFanNPUMax{}\,$^\circ$C and the CPU at \nCostNoFanCPUMax{} instead of
\nCostNoFanFanCPUMax{}\,$^\circ$C. The rate sessions ran at a mean battery voltage of \nCostVoltTwo{}\,V (2B upgrade)
and \nCostVoltFour{}\,V (4B) on the OnePlus 15 and \nCostThirteenVolt{}\,V on the 13T. The end-to-end migration and
retraining ran at \nCostEndVolt{} and \nCostRetrainEndVolt{}\,V. Each row of Table~\ref{tab:cost} converts energy and
charge at its own session's voltage, so kJ and battery share are not in one fixed ratio across rows.

\textbf{Engine and software.} The engine trains on the NPU because, on its own benchmark batch, its fused CPU path is
about \nCPUPathSlowdown{} times slower~\cite{npu-training}. The frozen weights are held on the NPU in 8-bit form
(Q8\_0), except Qwen3.5-4B's linear-attention projections, which stay in FP16. The linear-attention recurrence runs in
the engine's fused CPU operators. In a short Qwen3.5-2B distillation run on the OnePlus 15 they raise throughput from
\nCostGDNUnfusedTwoKDTokPerSec{} tokens per second (reference PyTorch implementation) to
\nCostGDNFusedTwoKDTokPerSec{}. The adapter and the optimizer run in PyTorch on 8 CPU threads, with micro-batches of
one prompt and 16 accumulation steps. Each update matches the GPU's batch of 16, at the same learning rate
($2\times10^{-4}$ for distillation) and number of epochs. The phone's teacher phase, like the GPU's, computes logits
only at the answer position. The engine's fused LoRA kernel requires a dropout rate strictly between 0 and 1, but our
adapters train without dropout, so their LoRA branch runs through PEFT. On the GPUs the pipeline runs on PyTorch 2.12,
Hugging Face Transformers 5.17 and PEFT 0.21, on NVIDIA RTX PRO 6000 GPUs with 96\,GB of memory, one or two processes
per GPU.

\textbf{The second phone.} 
Table~\ref{tab:cost13t} shows that on the OnePlus 13T, the 2B upgrade’s migration from public or self-generated prompts takes 66 and 72 minutes at a fixed four epochs.
The fixed 4 epochs change accuracy by \ntwoPublicFourMinusSel{} and \ntwoSelfFourMinusSel{} points against
the selected length. With the selection run, the public-prompt migration takes \nCostPublicSelTwoHours{} hours. For
public prompts, held-out selection picks \nCostPublicSelFourEpochs{} epochs on the 4B upgrade and
\nCostPublicSelTwoEpochs{} on the 2B.

\begin{table}[t]
\caption{Migration costs on the OnePlus 13T for the 2B upgrade, using the same user and measurement protocol as Table~\ref{tab:cost}. Time is measured while charging. Energy is estimated from battery measurements with idle consumption excluded. $\Delta$ is the GPU accuracy difference between fixed and selected training lengths.}
\label{tab:cost13t}
\footnotesize
\setlength{\tabcolsep}{4pt}
\begin{tabular}{@{}lcccccccc@{}}
\toprule
 & \multicolumn{3}{c}{Minutes} & Tokens/s & \multicolumn{2}{c}{Energy} & Peak memory & \\
\cmidrule(lr){2-4}\cmidrule(lr){6-7}
Method & Phase 1 & Phase 2 & Total & phase 2 & kJ & \% battery & (GB) & $\Delta$ \\
\midrule
\multicolumn{9}{@{}l}{\emph{2B upgrade, Qwen3-1.7B $\rightarrow$ Qwen3.5-2B, OnePlus 13T}} \\
Stored inputs, selected length & 1.1 & 70.9 & 72.0 & 78 & 58.7 & 91 & 2.6 & -- \\
Stored inputs, 8 epochs & 1.1 & 19.7 & 20.8 & 78 & 16.7 & 26 & 2.6 & +0.1 \\
Retraining with labels & -- & 20.6 & 20.6 & 77 & 16.8 & 26 & 2.6 & -- \\
Public prompts, 4 epochs & 5.6 & 59.9 & 65.5 & 81 & 54.6 & 84 & 2.8 & \textminus 0.3 \\
Self-generated prompts, 4 epochs & 6.1 & 65.4 & 71.5 & 83 & 60.8 & 94 & 2.6 & 0.0 \\
Public prompts, selected length & 5.6 & 261.5 & 267.0 & 79 & 220.1 & 339 & 2.8 & -- \\
\bottomrule
\end{tabular}

\end{table}

\section{Phone–GPU training consistency}
\label{sec:app-equiv}

Phase~1 on the phone matches the GPU teacher's label probabilities to a mean KL divergence of at most
\nEquivTeacherKL{} per teacher. The teachers are the \nEquivTeacherUsers{} old personal models run on the phones,
\nEquivTeacherTwoUsers{} on the 2B upgrade and \nEquivTeacherFourUsers{} on the 4B. The top label is the same on
\nEquivTeacherArgmaxStoredSame{} of their \nEquivTeacherArgmaxStoredTotal{} stored inputs. On the 512 public and 512
self-generated prompts, run for \nEquivTeacherArgmaxPublicUsers{} of these teachers, the top label agrees on at least
\nEquivTeacherArgmaxPublic{} and \nEquivTeacherArgmaxSynth{} of 512. The phone's selection run picks the GPU's length
in \nEquivSelSameLength{} of \nEquivSelRuns{} selections. The other \nEquivSelOtherAdapters{}, both on public prompts,
match the GPU's checkpoint accuracy at the length the phone picked.

Phase~2 is compared item by item where the GPU run kept per-item predictions. This covers \nEquivAdapters{} adapters
trained on the phones at the method's selected length or with the retraining recipe, each user, fold and recipe
counted once. Of these, \nEquivIdentical{} predict every test item exactly as the GPU run of the same user and fold.
Over all of these, \nEquivItemsSame{} of \nEquivItemsTotal{} items agree. Of the \nEquivFixedAdapters{} distillations at a fixed
length, for which the GPU kept only checkpoint accuracy, \nEquivFixedExact{} match it exactly. The two phones, with
different NPUs, produce bit-identical adapters in all \nEquivBothPhonesAdapters{} configurations run on both
(\nEquivBothPhonesUsers{} users), and \nEquivRerunRuns{} reruns of \nEquivRerunGroups{} configurations on one phone are
bit-identical, so the differences are systematic, not noise. The largest share of differing items is for the user with
the fewest stored inputs (\nEquivWorstPrompts{}), whose \nEquivWorstEpochs{}-epoch distillation differs from the GPU on
\nEquivWorstDiff{} of \nEquivWorstItems{} items, \nEquivWorstNet{} net (McNemar $p=\nEquivWorstMcNemar{}$). For this
user, the phone's student scores the same when evaluated on the GPU, and its teacher outputs trained on the GPU give
the GPU result, so what remains is the NPU's training arithmetic. A GPU rerun of the same configuration on another machine differs from the original on
\nEquivWorstGpuRerunDiff{} items. All accuracy in this paper is from the GPU runs, and the phone's adapters are the cost
measurement.

\section*{Generative AI Usage Disclosure}
We used ChatGPT to improve language clarity and polish sentences. We also used Claude Code to assist with developing LoRA fine-tuning code and resolving runtime errors. All AI-assisted code was manually reviewed and revised by the authors, relevant functions were tested, and the corresponding experiments were rerun. The authors retain full responsibility for the technical contributions, experimental results, and final manuscript.

\clearpage

\end{document}